# Do ECG Foundation Models Transfer to Rare Cardiac Diseases? Evidence from Brugada Syndrome Detection

Beatrice Zanchi[a,b,*], Giuliana Monachino[a], Alvise Dei Rossi[a,c], Luigi Fiorillo[d,e], Georgia Sarquella-Brugada[f], Giulio Conte[g,h] and Francesca Dalia Faraci[a]

[a] *University of Applied Sciences and Arts of Southern Switzerland, Department of Innovative Technologies, Via La Santa 1, Lugano, 6900, Ticino, Switzerland*

[b] *University of Zurich, Department of Quantitative Biomedicine, Winterthurerstrasse 190, Zurich, 8057, Zurich, Switzerland*

[c] *Università della Svizzera Italiana, Faculty of Informatics, Via Giuseppe Buffi 13, Lugano, 6900, Ticino, Switzerland*

[d] *Neurocenter of Southern Switzerland, Ente Ospedaliero Cantonale, Movement Disorder Research Group, Neurology Department, Via Tesserete 46, Lugano, 6900, Ticino, Switzerland*

[e] *Sleep Research Institute, C. del Padre Damián 44, Chamartín, Madrid, 28036, Spain*

[f] *Hospital Sant Joan de Déu, Inherited Cardiac Diseases and Sudden Death Unit, Pg. de Sant Joan de Déu 2, Esplugues de Llobregat, 08950, Barcelona, Spain*

[g] *Cardiocentro Ticino Institute, Ente Ospedaliero Cantonale, Division of Cardiology, Via Tesserete 48, Lugano, 6900, Ticino, Switzerland*

[h] *Università della Svizzera Italiana, Faculty of Biomedical Sciences, Via Giuseppe Buffi 13, Lugano, 6900, Ticino, Switzerland*

ABSTRACT

Background: Foundation models (FMs) trained on large-scale unlabeled physiological data have emerged as a promising paradigm for medical artificial intelligence. However, their ability to capture clinically meaningful and transferable representations for rare diseases remains largely unproven. This study investigates whether FM pre-training provides genuine clinical generalization benefits beyond improved optimization in the context of rare electrocardiographic (ECG) phenotypes.

Methods:We systematically evaluated nine publicly available ECG FMs for Brugada syndrome detection on BrSwiss cohort (294 patients, 87 cases) and the independent external HUCA cohort (363 patients, 76 cases). Each FM was evaluated under three strategies (from-scratch training, linear probing, full fine-tuning) across several experimental configurations, including a 3% data ablation and zero-shot cross-site transfers.

Results: Pre-training was necessary for high-capacity architectures unable to converge from scratch ($\Delta$AUC up to +0.411, $p < 0.05$), but did not provide a significant gain for compact architectures that already converged on the labeled data alone. On the full BrSwiss dataset, the best fine-tuned FM (ECG-CPC, AUC = 0.962) only marginally exceeded the strongest supervised baseline (ECG-CPC from scratch, AUC = 0.932; $p = 0.091$). At matched training-set size, the FMs' data-efficiency advantage observed on BrSwiss-3% ($\Delta$AUC = +0.055, $p < 0.01$) did not replicate on HUCA . Under zero-shot cross-site transfer, FM-based pipelines did not generalise better than supervised baselines, with all strategies approaching chance-level performance.

Conclusion: For Brugada syndrome detection, FM pre-training is mechanical rather than semantic, providing optimization stability rather than transferable clinical knowledge. These findings challenge the assumption that large-scale pre-training inherently encodes clinically meaningful representations, highlighting the central role of model architecture and data-domain alignment.

## 1. Introduction

Deep learning has significantly advanced the automated analysis of electrocardiograms (ECGs) in the diagnosis of common cardiac conditions [1, 2, 3, 4]. However, the diagnosis of rare cardiac diseases remains a significant challenge. Standard supervised learning approaches rely heavily on large-scale, high-quality labeled datasets. Due to the inherent low prevalence of these rare conditions, such datasets are scarce, often leading to overfitting and poor generalization when applying traditional supervised techniques [5, 6, 7].

*Corresponding author

beatrice.zanchi@supsi.ch (B. Zanchi)
ORCID(s): 0009-0009-3017-0645 (B. Zanchi)

The recent emergence of Foundation Models (FMs) introduces a new paradigm in ECG analysis. By leveraging vast repositories of unlabeled data to learn generalizable representations, these models offer a promising approach to mitigate the reliance on extensive annotated datasets [8]. The ECG FM landscape has rapidly diversified, producing a heterogeneous ecosystem of models that differ not only in architecture and self-supervised learning objective, but also in pre-training data, input specifications, training methodologies and intended deployment context [9, 10, 11, 12, 13, 14, 15]. Theoretically, the pre-trained knowledge embedded in FMs should provide a distinct advantage in low-resource settings, potentially bridging the gap where labeled data is insufficient. Rare cardiac diseases represent precisely this scenario. Their low prevalence makes large-scale labeled dataset collection inherently unfeasible, yet the clinical stakes of missed or delayed diagnosis are high. Under the theoretical premise that large-scale pre-training produces general-purpose representations transferable to downstream tasks with limited supervision [16], rare diseases should represent one of the most natural use cases for ECG FMs. Whether this theoretical advantage materializes in practice, however, remains poorly quantified. Existing FM benchmarks have focused predominantly on common cardiac conditions where labeled data is abundant [17, 18, 19, 20], leaving a critical evidence gap for the very settings where FM adoption would provide the greatest clinical value.

In this work, we address this gap by investigating whether ECG FMs, deployed under their native specifications, can provide an effective starting point for the detection of rare cardiac diseases. We focus on Brugada syndrome (BrS), an inherited arrhythmogenic condition with limited labeled data availability, as a representative instance of the low-data, high-stakes regime that characterizes rare disease diagnosis. Rather than isolating individual architectural components, we treat each FM as a complete deployable package (encompassing architecture, pre-training objective, input specifications, and learned representations) and assess their practical utility under realistic data constraints. Extending the benchmarking framework proposed in [17], our study evaluates a broad suite of publicly available FMs: ECGFounder [9], ECG-JEPA [10], ST-MEM [11], MERL-ViT and MERL-ResNet [12], ECGFM-KED [13], HuBERTECG [14], ECG-FM [15], and ECG-CPC [17].

The current analysis is structured around three interrelated research questions. First, we investigate the *source of discriminative advantage* in FM-based pipelines. Specifically, we disentangle the contribution of pre-trained representations from that of the underlying architecture by comparing fine-tuned FMs against their randomly initialized counterparts trained from scratch. This decomposition clarifies whether the observed performances are due to knowledge transferred from large-scale pre-training or merely to the capacity and inductive biases of the architecture itself. Second, we examine the *interaction between pre-training and training-set size*. Under the hypothesis that FM pre-training provides a stronger advantage when labeled data is highly scarce, we reduce the amount of available training data and characterize the scaling behaviour of both FM-based and supervised approaches. This allows us to identify the data regime in which pre-trained representations yield a meaningful benefit. Third, we assess the *cross-site generalization* of the evaluated approaches. Rare disease cohorts are inevitably drawn from specialized centers, introducing site-specific acquisition protocols and population characteristics that may limit external validity. By evaluating all models on an independent open-access cohort collected under different conditions, we quantify the degree to which performance degrades across sites and determine whether FM pre-training on diverse, multi-source data can confer greater robustness to distributional shift than supervised models trained exclusively on single-center data.

Our main contributions are summarized as follows:

- **Comprehensive FM benchmarking for rare cardiac disease detection.** We conduct a systematic evaluation of nine publicly available ECG FMs for Brugada syndrome detection, assessing each as a complete deployable package. The evaluation spans both an internal clinical dataset and an independent open-access cohort for external validation, providing practical investigation of whether FM characteristics (including architecture, pre-training paradigm, and input specifications) are associated with effective adaptation and generalization in low-data settings.

- **Disentangling the impact of pre-training.** Through controlled comparisons between fine-tuned foundation models and identically architected models trained from random initialization, we quantify the contribution of pre-training to downstream diagnostic performance.

- **Characterization of data efficiency and cross-site robustness.** We systematically evaluate the interaction between pre-training and training set size, identifying the data regime in which FMs provide a meaningful advantage, and assess cross-site generalization to determine if FM pre-training can confer greater robustness to distributional shift than single-center supervised training.

## 2. Related Works

**ECG Foundation Models benchmarking.** The rapid proliferation of ECG FMs has motivated a parallel line of research aimed at establishing rigorous, comparable evaluation protocols across this heterogeneous landscape. Wan et al. [18] introduced OpenECG, a benchmark assembling roughly 1.2 million 12-lead recordings from nine public centers to systematically contrast three self-supervised paradigms (SimCLR, BYOL, and MAE) across ResNet-50 and Vision Transformer backbones. The authors reported that feature-consistency (BYOL) and generative (MAE) objectives outperform contrastive learning and that publicly curated corpora can compete with proprietary ones for FM pre-training. Xu et al. [19] adopted a complementary perspective, interrogating the practical utility of FMs on the MIMIC-IV-ECG dataset by comparing language, general time-series, and ECG-specific FMs against time-series deep learning baselines on five tasks (RR-interval estimation, age, gender, potassium abnormality, and arrhythmia detection). Their findings suggested that general time series FMs and ECG-specific FMs reach a top-performance rate of approximately 80%, while exposing residual limitations on highly imbalanced clinical targets. Lunelli et al. [20] proposed BenchECG, a standardized suite spanning eight public datasets and ten heterogeneous tasks (classification, regression, detection, survival, and segmentation). Among recently published analyses, Al-Masud et al. [17] benchmarked eight publicly available ECG FMs on 26 clinically relevant tasks across 12 public datasets. Their analysis showed that FMs deliver consistent gains primarily in adult ECG interpretation and improve label efficiency by 3.3 to 9 times over supervised baselines. Collectively, these benchmarks have established credible evaluation protocols, yet they remain confined to common cardiac conditions and abundantly labeled data, leaving unexplored the low-prevalence, high-stakes regime of rare cardiac diseases.

**Foundation Models for rare cardiac conditions** A growing body of work has begun probing FM-based transfer learning in low-prevalence cardiac settings. Hu et al. [21] pre-trained a convolutional FM on 1.6M cardiologist-labeled ECGs covering 68 common diagnoses and fine-tuned it on three rare targets (carcinoid syndrome, pericardial constriction, rheumatic doming), outperforming a single DNN trained from scratch on each task. Ronan et al. [22] pre-trained a VICReg encoder on 1.2M unlabeled ECGs and fine-tuned it for Brugada syndrome, surpassing a single CNN trained from scratch (AUROC 0.88 vs. 0.76) and uncovering mislabelled cases that revised the institutional prevalence. Both works, however, validate a single custom FM against a single trained-from-scratch baseline, leaving open which characteristics of the broader public FM ecosystem transfer to the rare-disease regime. This is the gap our analysis addresses.

## 3. Methods

### 3.1. Study design and population

The internal dataset (hereafter *BrSwiss*, for Brugada Syndrome Swiss cohort) was collected at a tertiary referral center in Switzerland under approved ethics protocols (Swiss Ethics, approval number: BASEC 2019-00754/CE 3476). It comprises 87 Brugada syndrome patients diagnosed according to current guidelines and the Shanghai Score System [23]. The positive class encompasses (i) spontaneous type 1 Brugada pattern, (ii) pharmacologically induced type 1 pattern following sodium channel blocker challenge (Ajmaline/Flecainide test), and (iii) fever-induced type 1 pattern. The control group comprises 207 adult patients with a negative Ajmaline provocation test result or other confirmed cardiac diagnoses unrelated to Brugada syndrome. In addition to the BrSwiss dataset, this study exploits *HUCA dataset*, collected retrospectively from patients evaluated at the Cardiology Department of the Hospital Universitario Central de Asturias (HUCA) [24]. The HUCA dataset comprises 363 adult subjects, including 76 patients diagnosed with Brugada syndrome and 287 control subjects. Diagnostic classification followed internationally accepted consensus criteria. The positive class encompasses (i) spontaneous type 1 Brugada pattern identified on the basal ECG, and (ii) pharmacologically induced type 1 pattern unmasked following provocative challenge (Flecainide test). Each recording consists of 12 standard ECG leads acquired over a 12-second window at a sampling frequency of 100 Hz, with signal amplitudes expressed in millivolts (mV) according to standard diagnostic ECG conventions.

All patients included in this study, from both the BrSwiss and HUCA datasets, are over 18 years of age. The BrSwiss dataset is not publicly available and was not used in the pretraining corpora of any of the FMs evaluated in this work. The HUCA dataset is publicly released [24], but its release postdates the pretraining of all FMs included in our benchmark, ensuring that no patient in either cohort was exposed to the models during their pretraining phase. Detailed data collection and preprocessing are disclosed in Appendix Section A.

### 3.2. Supervised baselines

To contextualize the performance of the available ECG FMs and to accurately isolate the impact of the pre-training phase, we considered a direct supervised baseline for every model evaluated in this study. Specifically, each FM architecture was also trained from scratch (i.e. using randomly initialized weights) exclusively on the labeled target datasets. Unlike the pre-trained FM instances, these supervised counterparts do not benefit from prior exposure to large-scale unlabeled ECG repositories; they rely entirely on the limited labeled data available from the BrSwiss and HUCA datasets. This one-to-one paired comparison provides a consistent reference point, ensuring that any observed performance gains can be attributed directly to the self-supervised pre-trained representations rather than to inherent architectural advantages. Consequently, our supervised baseline suite perfectly mirrors the architectural diversity of the evaluated FMs, encompassing state-space models (such as S4, the backbone of ECG-CPC [25]), convolutional neural networks (such as ResNet-derived models, exploited by MERL-ResNet or ECGFM-KED [26]), and various Transformer-based architectures (such as HuBERT for HuBERTECG [27]). All supervised baselines are trained from randomly initialized weights using Binary Cross-Entropy loss under identical experimental conditions with respect to their counterparts, thereby yielding a coherent and fair benchmark for investigating the contribution of self-supervised pre-training on overall performance.

### 3.3. Foundation models

In the main corpus of our analysis, we evaluate nine FMs: ECGFounder [9], ECG-JEPA [10], ST-MEM [11], MERL (ViT and ResNet backbones), [12], ECGFM-KED [13], HuBERTECG [14], ECG-FM [15] and ECG-CPC [17]. The selection criteria were two-fold: we prioritize models with publicly available weights and trained on 12-leads ECG data. At the same time, we ensure coverage of the diverse landscape of ECG FM approaches currently available, spanning a range of architectural priors (Transformer-based, Convolutional, and Structured State-Space Models) and self-supervised learning objectives (e.g. masked modeling, contrastive learning). The heterogeneity in input specifications across models, including window duration (2.4s to 10s), sampling rate (100 Hz to 500 Hz), and parameter count (2.9M to 94.6M), reflects the real-world diversity of the current ECG FM ecosystem and is an inherent property of the pipeline being evaluated. Details about each FM package are summarized in Table 1.

**Transformer-based models**. A significant portion of the benchmarked FMs leverages the Transformer architecture [28]. ECG-FM employs a wav2vec 2.0-style backbone [29] with a hybrid Self-Supervised Learning (SSL) approach combining signal masking and contrastive learning (CMSC) [30]. ECG-JEPA utilizes a Vision Transformer (ViT) [31] within a Joint-Embedding Predictive Architecture (JEPA) [32], optimized via a Smooth L1 loss in the latent space. ST-MEM adopts a ViT backbone trained as a Masked Autoencoder (MAE) [33]. HuBERTECG follows the HuBERT framework [27] for masked prediction of hidden units, while MERL-ViT [12] uses a ViT backbone in a multimodal (text-signal) contrastive setup.

**CNN and SSM-based models**: ECG-CPC is built on SSM [34] and trained with Contrastive Predictive Coding (CPC) [35]. ECGFounder uses a RegNet/Net1D CNN backbone [36, 37], pre-trained through supervised multi-label classification using a specialized Positive Unlabeled (PU) Learning loss [38]. ECGFM-KED and MERL-ResNet represent multimodal approaches: the former uses an XResNet101 backbone [39] with signal-text alignment, while the latter utilizes a ResNet18 [26].

### 3.4. Experiments and training protocol

We evaluate the models under three distinct experimental settings: *Train from Scratch (FS)*, *Linear Probing (LP)*, and *Full Fine-tuning (FT)*, as shown in Figure 1. In the *Train from Scratch* setting, the model architectures are initialized with random weights and trained entirely on the target dataset (learning rate of $\eta = 10^{-3}$), serving as our fully supervised baselines. In the *Linear Probing* setting, the pre-trained encoders remain frozen, acting as static feature extractors, while only the classification head is trained with a learning rate of $\eta = 10^{-2}$. Finally, for *Full Fine-tuning*, the entire network is initialized with pre-trained weights and optimized using a Layer-wise Learning Rate Decay (LLRD) scheme [40]. In this setting, the model is divided into three depth regions: the classification head is updated at $\eta = 10^{-3}$, the medium-depth layers at $\eta = 10^{-4}$, and the initial layers at $\eta = 10^{-5}$ to preserve the low-level pre-trained representations. Further details can be found in Appendix Section B. All models across the three settings are optimized using the Adam optimizer [41] by minimizing the Binary Cross-Entropy loss. To prevent overfitting and ensure a fair

**Table 1**
Summary of the benchmarked Foundation Models.

| Model Name | Backbone Architecture | Learning Framework | ECG Samples | Input Specs | Params |
|---|---|---|---|---|---|
| ECG-CPC | SSM | SSL (CPC) | 10.7M[a] | 2.5s @ 240 Hz | 2.9 M |
| ECG-FM | Transformer (wav2vec 2.0) | Hybrid SSL (CMSC) | 1.5M[b] | 5s @ 500 Hz | 92.1 M |
| ECG-JEPA | Transformer (ViT) | SSL (JEPA) | 174k[c] | 10s @ 250 Hz | 87.1 M |
| ECGFM-KED | CNN (XResNet101) | Multimodal (Signal-Text) | 800k[d] | 10s @ 500 Hz | 9.6 M |
| ECGFounder | CNN (RegNet / Net1D) | Supervised (Multi-label) | 10.7M[a] | 2.5s @ 500 Hz | 33.8 M |
| HuBERTECG | Transformer (HuBERT) | SSL (Masked Prediction) | 9.1M[e] | 5s @ 100 Hz | 94.6 M |
| MERL-ResNet | CNN (ResNet18) | Multimodal SSL | 800k[d] | 2.5s @ 500 Hz | 4.6 M |
| MERL-ViT | Transformer (ViT) | Multimodal SSL | 800K[d] | 10s @ 500 Hz | 5.6 M |
| ST-MEM | Transformer (ViT) | SSL (MAE) | 174k[c] | 2.4s @ 250 Hz | 90.3 M |

SSM: Structured State-Space Model; CPC: Contrastive Predictive Coding; CMSC: Contrastive Multi-Segment Coding; ViT: Vision Transformer, JEPA: Joint-Embedding Predictive Architecture; MAE: Masked Autoencoder;
**a**: HEEDB; **b**: CPSC2018, CPSC-Extra, PTB-XL, Georgia, Ningbo, Chapman, MIMIC-IV-ECG; **c**: Chapman, Ningbo, CODE-15; **d**: MIMIC-IV-ECG; **e**: CODE, PTB, CPSC2018, CPSC-Extra, PTB-XL, Georgia, Chapman, Ningbo, Hefei, SPH, MIMIC-IV-ECG.

termination criterion across FMs operating under heterogeneous input specifications, we employ early stopping based on validation AUC with a patience of 15 epochs ($\delta = 0.001$) and a maximum budget of 200 epochs. Importantly, early stopping decouples the effective training duration from the nominal epoch count. Thus, models processing shorter input windows (and therefore producing more segments per ECG recording) do not systematically benefit from a fixed epoch budget.

**Data efficiency and scaling behaviour.** A core motivation for adopting foundation models in rare-disease research is the expectation that pre-trained representations should yield their largest benefit when labeled data are scarce. To test this hypothesis directly, we designed an ablation study that places FM-based and supervised approaches under controlled conditions of data scarcity. We reduced the number of ECG windows per patient rather than the number of patients themselves. An excessive reduction in patient count would expose all models to severe overfitting on a handful of subjects, producing performance estimates dominated by inter-patient variability rather than by the data-efficiency behavior we aim to characterize. The reduction ratio was set to balance two practical considerations. First, the BrSwiss cohort, while small by general AI standards, is still larger than the smallest cohorts encountered in rare-disease practice. A substantial reduction is therefore necessary to reach the regime in which the data-efficiency claim of FMs is most meaningful. Second, we sought a training-set size comparable to that of the external HUCA cohort, enabling a downstream comparison on equal footing in terms of data volume. A ratio of 3% of the original training windows per patient (retained in both training and validation sets) satisfies both criteria. The test set remains fixed and identical to that of the full-scale experiments, and all experiments follow the same three-fold cross-validation protocol described above. This reduced configuration (hereafter BrSwiss-3%) is compared with two reference settings: (i) the full-scale BrSwiss-100% experiment, which varies only the amount of training data and thereby isolates the effect of data scarcity on each adaptation strategy; and (ii) the HUCA dataset, which has a naturally comparable training-set size but originates from a different clinical center and country, probing whether the data-efficiency advantages observed on the BrSwiss cohort replicate on an independent cohort of similar size.

**Cross-site generalization.** Beyond data efficiency, a second open question is whether large-scale, multi-source pretraining equips foundation models with representations that transfer robustly across clinical sites. To assess the external validity of the evaluated approaches, we perform a series of cross-site experiments in which models trained on one dataset are directly applied to the other without any additional fine-tuning. This zero-shot transfer protocol isolates the effect of distributional shift arising from differences in acquisition hardware, clinical protocols, and population characteristics between the two centers. Three configurations were evaluated, each designed to probe a distinct facet of cross-site behaviour: (i) models trained on the full BrSwiss dataset (BrSwiss-100%) and tested on HUCA, which assesses whether a model developed on a moderately sized single-center cohort transfers to an unseen center; (ii) models trained on the reduced BrSwiss dataset (BrSwiss-3%) and tested on HUCA, which characterizes the joint effect of data scarcity and distributional shift, the most challenging combination in rare-disease practice; and (iii) models trained on HUCA and tested on the BrSwiss dataset, which verifies whether the patterns observed in the first two transfers depend on the choice of source cohort or generalize in both directions. All three strategies (FS, LP, and FT) are evaluated in

**Table 2**
BrSwiss Population Characteristics.

| | BrS (n=87) | Controls (n=207) |
|---|---|---|
| ***Demographic Characteristics*** | | |
| Male Sex, n (%) | 63 (72.4) | 156 (75.4) |
| Age (years) | 53 (43–64)* | 69 (62–74) |
| ***Clinical Characteristics — BrS*** | | |
| Spontaneous Type I Pattern, n (%) | 24 (27.5) | – |
| Ajmaline-induced Type I Pattern, n (%) | 55 (63.2) | – |
| Fever-induced Type I Pattern, n (%) | 8 (9.2) | – |
| *SCN5A* pathogenic variant, n (%) | 10 (11.5) | – |

Continuous data are represented as median ($25^{th}$ percentile - $75^{th}$ percentile).
Categorical data are represented as number (percentage).
* $p < 0.01$ – BrS vs Controls;

**Table 3**
Segment counts for the training and testing subsets across different input durations.

| | BrSwiss Dataset (N=294) | | | HUCA Dataset (N=363) | |
|---|---|---|---|---|---|
| Input Size (s) | Train | 3% Train | Test | Train | Test |
| 2.4[a] | 23'522 | 970 | 12'132 | 1'210 | 605 |
| 2.5[b] | 22'581 | 776 | 11'646 | 968 | 484 |
| 5.0[c] | 11'267 | 388 | 5'812 | 484 | 242 |
| 10.0[d] | 5'612 | 194 | 2'897 | 242 | 121 |

Segment counts are identical across models sharing the same input size. [a]: ST-MEM; [b]: ECG-CPC, ECGFounder, MERL-ResNet; [c]: ECG-FM, HuBERTECG; [d]: ECG-JEPA, ECGFM-KED, MERL-ViT.

each configuration, so that the effect of pre-training on cross-site robustness can be assessed at every level of model adaptation. By comparing cross-site performance across these settings, we quantify the degree to which performance degrades when moving between sites and determine if FM pre-training on large-scale, multi-source data can confer greater robustness to distributional shift with respect to supervised models trained exclusively on single-center data.

Evaluation protocols and statistical analysis are detailed in Appendix Section C.

# 4. Results

## 4.1. Population and dataset characteristics

**Population characteristics.** Table 2 summarizes the demographic and clinical characteristics of the BrSwiss dataset. BrS patients show a similar gender distribution compared to controls, with a slight male predominance across all groups (BrS: 72.4%, controls: 75.4%; $p = 0.70$). Notably, Brugada cohort is significantly younger than controls at the time of ECG recording (BrS: median 53 years [IQR: 43–64]; controls: median 69 years [IQR: 62–74]; $p < 0.01$). Within the BrS cohort, the majority of patients presented with an Ajmaline-induced Type I ECG pattern (N=55, 63.2%). Genetic analysis performed on all Brugada subjects revealed that 11.5% patients carried pathogenic or likely pathogenic variants in the *SCN5A* gene. Detailed HUCA population characteristics can be found at [24].

**Datasets characteristics.** Table 3 summarizes the number of ECG segments generated per model across training (including validation) and test splits for both datasets. Segment counts vary substantially across FM packages as a direct consequence of their native input window durations (from 2.4s to 10s). The resulting training sets span from approximately 5,600 to 23,500 segments for the full BrSwiss dataset, and from 242 to 1,210 for HUCA.

## 4.2. Foundation models evaluation

**Pre-training contribution.** Table 4 reports patient-level AUC across the three experimental settings on the full BrSwiss dataset. We structure this analysis along two complementary axes: a *within-model* comparison, which contrasts

**Table 4**
Patient-level AUC (mean [95% CI]) on the BrSwiss-100% dataset across three experimental settings.

| Model | FS | LP | FT |
|---|---|---|---|
| HuBERTECG (94.6 M) | 0.510 [0.445–0.589] | 0.864*[0.812–0.912] | 0.887*[0.847–0.925] |
| ECG-FM (92.1 M) | 0.503 [0.424–0.555] | 0.852*[0.804–0.894] | 0.914*[0.882–0.941] |
| ST-MEM (90.3 M) | 0.785 [0.739–0.833] | 0.926*[0.899–0.949] | 0.928*[0.903–0.952] |
| ECG-JEPA (87.1 M) | 0.912 [0.872–0.947] | 0.873*[0.826–0.909] | 0.943 [0.917–0.969] |
| ECGFounder (33.8 M) | 0.904 [0.875–0.938] | **0.955***[0.930–0.973] | 0.890 [0.845–0.937] |
| ECGFM-KED (9.6 M) | 0.880 [0.819–0.933] | 0.745*[0.699–0.802] | 0.921 [0.879–0.951] |
| MERL-ViT (5.6 M) | 0.861 [0.812–0.900] | 0.800*[0.737–0.850] | 0.759*[0.702–0.827] |
| MERL-ResNet (4.6 M) | 0.913 [0.868–0.951] | 0.608*[0.537–0.676] | 0.951*[0.930–0.976] |
| ECG-CPC (2.9 M) | **0.932** [0.901–0.961] | 0.931 [0.904–0.957] | **0.962** [0.947–0.978] |

**Training strategies**: FS = Training from Scratch; LP = Linear Probing; FT = Full Fine-Tuning.
*****: Statistically significant difference compared to the supervised counterpart trained from scratch (DeLong's test, FDR-adjusted $p < 0.05$).

the three adaptation strategies (FS - Training from scratch, LP - Linear Probing, FT - Full Fine-Tuning) of each architecture to isolate the contribution of pre-trained weights, and a *between-model* comparison, which contrasts the best pre-trained pipeline against the strongest supervised baseline to assess whether pre-training yields a practical advantage at the level of the best available model.

The supervised setting (FS) serves as a controlled baseline to assess each architecture's intrinsic capability to learn BrS-discriminative features without pre-trained initialization. Compact architectures such as ECG-CPC (AUC = 0.932 [0.901-0.961]) and MERL-ResNet (AUC = 0.913 [0.868-0.951]) already achieve strong performance when trained from scratch. Conversely, the largest transformer models (e.g. HuBERTECG, ECG-FM, ST-MEM) fail to converge (AUC $\approx$ 0.5), consistent with the challenge of training high-capacity models on small medical datasets [42, 43].

Within each architecture, the comparison between FT and FS isolates the contribution of pre-trained weights from that of architectural capacity. The effect of pre-training is highly heterogeneous across FMs. For those models whose architectures fail to converge when randomly initialized (e.g. HuBERTECG $AUC_{FS}$ = 0.510; ECG-FM $AUC_{FS}$ = 0.503) pre-training is essential, yielding large and statistically significant gains under fine-tuning ($\Delta$AUC = +0.377 and +0.411, respectively; $p < 0.05$). ST-MEM exhibits a similar, though more moderate, pattern ($\Delta$AUC = +0.143, $p < 0.05$). By contrast, for models that already achieve strong FS performance (e.g. ECG-CPC AUC = 0.932; MERL-ResNet AUC = 0.913; ECG-JEPA AUC = 0.912; ECGFounder AUC = 0.904; and ECGFM-KED AUC = 0.880) the marginal gain from pre-training is modest and not statistically significant. Notably, ECGFounder and MERL-ViT exhibit a negative fine-tuning variation in performance ($AUC_{FT} < AUC_{FS}$).

Within each architecture, the comparison between LP and FT characterizes the quality of pre-trained representations for BrS detection. For the majority of models (seven out of nine FMs), FT outperforms LP, indicating that pre-trained features require task-specific adaptation through deeper network layers to capture BrS-discriminative patterns. This gap is most pronounced for models trained with contrastive or multimodal objectives (MERL-ResNet: $AUC_{LP}$ = 0.608 vs $AUC_{FT}$ = 0.951; ECGFM-KED: $AUC_{LP}$ = 0.745 vs $AUC_{FT}$ = 0.921). A notable exception is ECGFounder, for which LP (0.955) outperforms FT (0.890). Unlike all other benchmarked FMs, ECGFounder was pre-trained through a supervised multi-label classification task, producing representations that may already encode clinically relevant cardiac features. Further fine-tuning on the small BrS dataset may degrade these through catastrophic forgetting.

To provide a more stringent test of the practical value of pre-training, we complement the within-model analysis with a between-model comparison, assessing whether the best FM-based pipeline outperforms the strongest supervised baseline overall. On BrSwiss-100%, the best FT model (ECG-CPC, AUC = 0.962) and the best LP model (ECG-Founder, AUC = 0.955) exceed the best from-scratch baseline (ECG-CPC, AUC = 0.932) by +0.031 and +0.023 AUC, respectively. However, neither improvement reaches statistical significance ($p$ = 0.091 and $p$ = 0.186). Thus, while pre-training yields significant within-model gains for several architectures, it does not provide a statistically detectable advantage at the level of the best available pipeline. Complete results tables and saliency maps analysis for this experiment can be found in the Appendix (Table 11, Figure 5).

**Data efficiency and scaling behaviour.** Figure 2 reports models' performance when the number of BrSwiss ECG windows per patient is reduced to 3% of the original count, while preserving the full patient cohort and the original test set. Complete results tables and saliency maps analysis for this experiment can be found in the Appendix. We first examine, within each architecture, how the three adaptation strategies respond to the reduction in training data. In the transition from the full training set to the reduced size, the average and median AUC values across the nine FMs decrease in all three settings, although with markedly different magnitudes. FS drops from a mean of 0.800 at 100% to 0.714 at 3% (median: 0.880 vs. 0.781), LP from 0.839 to 0.804 (median: 0.864 vs. 0.845), and FT from 0.906 to 0.869 (median: 0.913 vs. 0.868). The corresponding mean degradations are −0.086 for FS, −0.035 for LP, and −0.037 for FT, with median degradations of −0.066, −0.019, and −0.038, respectively. The substantially larger drop observed in the FS setting is expected. Indeed, high-capacity architectures are known to underperform when trained from scratch on small datasets, and three out of the nine FS models (ST-MEM, ECGFounder, MERL-ViT) lose more than 0.10 AUC, with ST-MEM exhibiting a drop of −0.233. More informative is the comparison between the relative degradations across strategies. While AUC decreases under all three regimes, the mean drop in FS is roughly twice that observed under LP or FT (−0.086 versus −0.035 and −0.037).

To verify that the trends observed on BrSwiss-3% are not specific to its cohort, we replicated the analysis on HUCA, which has a comparable training-set size (see Table 3). While absolute AUC values cannot be directly compared across cohorts, the hierarchy across training strategies is preserved. As shown in Figure 3 (and in the complete Appendix Table 7, 13), mean AUC follows the same ordering, with FS being the weakest (0.675), LP intermediate (0.805) and FT the strongest (0.835). Across both datasets, the same three FMs (ECG-CPC, MERL-ResNet, and ECGFounder) emerge as the most robust. They are the only models achieving FS AUC $\geq$ 0.85 on BrSwiss-3% and HUCA, and MERL-ResNet reaches the highest FT AUC on both (0.946 and 0.906, respectively). However, a clear divergence appears in the within-model comparison of the statistical contribution of pre-training. On BrSwiss-3%, ECG-CPC (FT), MERL-ResNet (LP and FT), and ECGFounder (LP and FT) all show significant gains over their FS baseline (DeLong's test, FDR-adjusted $p < 0.05$). On HUCA, none of these models show a statistically significant difference between FS and either LP or FT ($p \geq 0.05$): ECGFounder (FS = 0.896 vs FT = 0.891), MERL-ResNet (FS = 0.857 vs FT = 0.906), and ECG-CPC (FS = 0.873 vs FT = 0.873). For the architectures that already converge from scratch, the benefit of pre-training is therefore not replicated on the HUCA cohort. Conversely, the large transformer models (HuBERTECG, ECG-FM, ECG-JEPA, ST-MEM) that fail to converge from scratch on both datasets recover competitive performance only after LP or FT, with statistically significant gains preserved across both cohorts.

As for BrSwiss-100%, we investigate the between-model comparison, assessing whether the best FM-based pipeline outperforms the strongest supervised baseline overall. On BrSwiss-3%, the best FT model (MERL-ResNet, AUC = 0.946) and the best LP model (ECGFounder, AUC = 0.944) exceed the best from-scratch baseline (ECG-CPC, AUC = 0.891) by +0.055 and +0.053 AUC, respectively, with both differences reaching statistical significance ($p = 0.0015$ and $p = 0.0079$). On HUCA, however, the gap closes: the best FT model (MERL-ResNet, AUC = 0.906) exceeds the best from-scratch baseline (ECGFounder, AUC = 0.896) by only +0.010 AUC ($p = 0.715$), and the best LP model (ST-MEM, AUC = 0.862) is below the supervised baseline by −0.033 AUC ($p = 0.277$).

**Cross-site generalization.** Figure 4 reports patient-level AUC under the three zero-shot cross-site experiments. Further results can be found in the Appendix. Across all configurations, performance degrades substantially relative to the within-site experiments: mean AUC across the nine models ranges from 0.504 to 0.584 depending on the strategy, and the large majority of models fall below 0.65 AUC regardless of adaptation strategy. In the most extreme case, HuBERTECG produces constant predictions on the HUCA dataset across all strategies (AUC = 0.500, with degenerate confidence intervals), indicating a complete failure of transfer.
We first assessed, within each architecture, whether adaptation of the pre-trained weights improves over the matched from-scratch baseline. Fine-tuning yields a statistically significant gain in only a minority of models (4/9, 3/9, and 3/9 for train BrSwiss-100%, BrSwiss-3%, and HUCA, respectively), and linear probing in even fewer cases (2/9, 1/9, and 2/9; DeLong's test, FDR-adjusted $p < 0.05$). We then reframed the comparison at the level of the best available pipeline, testing each FM against the single strongest from-scratch baseline of the same configuration. Under this more stringent criterion, the advantage of pre-training largely disappears. Training on BrSwiss-100% and testing on HUCA, only ECG-CPC (FT, AUC = 0.724) significantly outperforms the best from-scratch baseline (ECG-JEPA, AUC = 0.620; $p < 0.05$). Training on BrSwiss-3% and testing on HUCA, only ECG-FM (FT = 0.710 and LP = 0.703) exceeds the best from-scratch model (ECG-JEPA, AUC = 0.565; $p < 0.05$). Training on HUCA and testing on BrSwiss, no FM significantly outperforms the strongest from-scratch baseline (ECG-CPC, AUC = 0.715); the closest case, ECGFM-KED (FT, AUC = 0.804), reaches only borderline significance ($p = 0.06$).

To assess whether the amount of source training data influences cross-site transfer, we compared the two configurations in which models were trained on the BrSwiss dataset and tested on HUCA, differing only in the size of the source training set (100% versus 3% of training windows). Under from-scratch training and full fine-tuning, the larger source set yielded a small improvement in transferred performance (mean $\Delta$AUC = +0.035 and +0.027 in favor of the 100% setting, with 7/9 and 6/9 models improving, respectively), consistent with the expectation that more source data produce more robust feature representations. Under linear probing, however, the effect was reversed: the 3% setting performed marginally better on average (mean $\Delta$AUC = −0.022; only 1/9 models improved with more data). A plausible explanation is that, with the encoder frozen, additional source data leads the linear classifier to specialize more closely on site-specific characteristics of the BrSwiss cohort, which in turn reduces its transferability to HUCA. In all cases, the magnitude of these effects ($\leq$ 0.035 AUC) was small relative to the overall performance drop induced by cross-site transfer, indicating that source training-set size is a second-order factor compared with the distributional shift between acquisition sites.

## 5. Discussion

The current study is designed to test whether ECG foundation models provide a practical advantage over supervised baselines for the detection of a rare inherited cardiac disease, Brugada syndrome, and to identify the conditions under which pre-training is most beneficial. Across within-model, between-model, and cross-site analyses, a coherent picture emerges, which we discuss along four main themes.

**The role of pre-training is mechanical rather than semantic.** Pre-training contributes to downstream performance primarily by stabilizing optimization for architectures that would otherwise fail to converge, rather than by transferring task-relevant cardiac representations. On the full BrSwiss dataset, large transformer models unable to converge from scratch (HuBERTECG, ECG-FM, ST-MEM) show large and significant gains under fine-tuning. On the other hand, in the architectures that already converge during supervised training (ECG-CPC, MERL-ResNet, ECG-JEPA, ECGFounder, ECGFM-KED) the gain is not statistically relevant. The same pattern becomes evident when the comparison is reframed at the level of the best pipeline rather than within each architecture (between-model analysis). On the full BrSwiss dataset, the highest AUC is achieved by the best fine-tuned FM (ECG-CPC, AUC = 0.962), with the best linear-probed FM (ECGFounder, AUC = 0.955) and the strongest supervised baseline (ECG-CPC, AUC = 0.932) reaching comparable values, neither difference being statistically significant. The replication on the HUCA cohort reinforces this picture, since the best FT model exceeds the best supervised baseline by only +0.010 AUC ($p = 0.715$), and the best LP model is actually below the supervised baseline (−0.033 AUC, $p = 0.277$). Across both cohorts, therefore, when the most competitive supervised model is selected as a comparator, the apparent advantage conferred by pre-training is markedly reduced and no longer statistically detectable. The strongest within-model gains observed in the results reflect the relative weakness of the matched FS counterpart rather than a genuine superiority of FM-based pipelines over the best supervised alternative. These observations are limited to a single rare condition evaluated on two cohorts, and we cannot exclude that broader pre-training data or other rare cardiovascular phenotypes may yield different patterns. Nonetheless, the convergence of the within-model and between-model analyses, replicated across two independent cohorts, calls for a reconsideration of the assumption that ECG FMs uniformly transfer clinically meaningful representations to rare disease tasks.

**The data-efficiency advantage of pre-training is cohort-specific.** The data efficiency analysis quantifies how pre-training interacts with training set size. Reducing the training set to 3% of the original size on the BrSwiss dataset produces a markedly asymmetric degradation across strategies, with mean AUC dropping by −0.086 under FS but only by −0.035 and −0.037 under LP and FT, respectively. The absolute performance of LP and FT remains essentially stable as data become scarcer, while the supervised baseline deteriorates substantially. In this regime, the between-model comparison also shows a statistically significant advantage of pre-training over the strongest trained from-scratch model (+0.055 and +0.053 AUC for FT and LP, $p < 0.01$). This advantage, however, does not replicate on the HUCA dataset, whose training-set size is comparable to BrSwiss-3%. The same between-model comparison yields no statistically significant difference, and the best supervised baseline (ECGFounder, AUC = 0.896) is essentially indistinguishable from the best FT model (MERL-ResNet, AUC = 0.906). The benefit of pre-training under data scarcity should therefore be interpreted with caution. This refines a common interpretation in the FM literature, where larger gains under data scarcity are often attributed to more effective knowledge transfer, and underscores the importance of further validation in claims of data-efficient learning.

**Pre-training does not confer cross-site robustness in rare disease detection.** The cross-site experiments (Figure 4) provide the most direct test of whether large-scale pre-training equips ECG FMs with representations that generalize beyond the cohort on which they are adapted. The results are unambiguous: every model and every adaptation strategy suffers a substantial loss of discriminative performance under zero-shot transfer between BrSwiss and HUCA, with most configurations approaching chance level. Within individual architectures, fine-tuning retains a statistically significant edge over training from scratch in a subset of models, indicating a modest residual benefit of pre-trained initialization under distributional shift. However, this benefit does not translate into a practical advantage at the level of the best available pipeline. Across all three cross-site experiments, at most a single FM significantly outperforms the strongest from-scratch baseline, and when testing HUCA-trained models on BrSwiss none does so, mirroring the within-site pattern. The choice between a fine-tuned FM and a supervised model trained from scratch therefore has little influence on cross-site generalization, since neither approach yields a model that transfers reliably across centers. The small gain afforded by larger source datasets under fine-tuning, and its absence under linear probing, further indicate that distributional shift between acquisition sites is the dominant obstacle. Achieving robust cross-site performance will likely require explicit strategies such as domain adaptation, rather than reliance on the implicit generalization assumed to emerge from large-scale pre-training.

**Architectural suitability is the most consistent predictor of performance.** Across both cohorts and all data regimes, the strongest performers are a small and stable set of architectures: ECG-CPC (2.9 M parameters), MERL-ResNet (4.6 M), and ECGFounder (33.8 M) consistently rank among the top three models on the BrSwiss and HUCA datasets. Crucially, these architectures lead the ranking even when trained from scratch. Indeed, on the full BrSwiss dataset, ECG-CPC (AUC = 0.932) and MERL-ResNet (AUC = 0.913) already outperform the best pre-trained large Transformer models, and on the HUCA cohort the strongest model across all strategies is the supervised baseline of ECGFounder (AUC = 0.896), not a pre-trained pipeline. Considered together, these results point to architecture, rather than the presence or modality of pre-training, as the primary determinant of which models perform best in this setting. Within the pre-trained models, our results suggest that performance is largely insensitive to parameter count. Either after linear probing or fine-tuning, large and compact FMs converge to comparable performance, indicating that pre-training tends to level the differences in raw capacity. This contrasts with the scaling behaviour documented in natural language processing [44, 45] and suggests that, in the low-data regime of rare disease detection, additional parameters confer little benefit once an adequate architecture and training procedure are in place. We note, however, that this observation likely depends on the absence (or near-absence) of Brugada syndrome from the pre-training corpora of the benchmarked FMs. When the downstream condition is well represented in the pre-training distribution, larger capacity may still translate into meaningful gains. The flattening of scaling we observe here should therefore be understood as specific to the regime in which the target task lies outside the effective coverage of the pre-training data.

**Clinical utility and translation.** Throughout this work, secondary metrics (Appendix Table 13–16) were reported at a fixed decision threshold of 0.5, which is appropriate for cross-model comparison but not for clinical use. In Brugada syndrome screening, the asymmetry of error costs (e.g., a missed case may translate into sudden cardiac death, whereas a false positive is typically resolved by a confirmatory Ajmaline challenge) calls for an operating point selected to maximize sensitivity at a clinically acceptable specificity. Even in our best within-site configuration (FT MERL-ResNet on BrSwiss-100%), achieving sensitivity above 0.90 would require explicit threshold tuning on a labeled validation cohort, and the resulting operating point should be revalidated whenever the deployment population or acquisition protocol changes. Beyond threshold calibration, the clinical translatability of the models deployed in this work will depend on three further elements. First, prospective validation in a true screening population (e.g. athletes, syncope clinics, or first-degree relatives of probands), where disease prevalence differs substantially from the case-control composition used here. Second, integration into the existing diagnostic workflow as a triage aid that flags suspicious ECGs for expert electrophysiological review, rather than as a stand-alone diagnostic instrument. Third, ongoing post-deployment monitoring of calibration and performance, particularly in light of the marked cross-site degradation we observed. As shown in our results (Figure 4), a model performing well at on center cannot be assumed to retain its operating characteristics elsewhere without explicit revalidation.

**Limitations.** First, evaluating each ECG FM under its native input specifications causes the number of segments per recording to vary across models, potentially confounding performance differences with the suitability of specific input windows. We mitigated this with validation-AUC-based early stopping, decoupling effective training duration from the nominal epoch count. Second, the two cohorts were acquired at different native rates (1000 Hz for BrSwiss, 100 Hz for HUCA), so most experiments combine down-sampling on one side with up-sampling on the other. Up-sampling cannot restore spectral content above the original Nyquist frequency, so FMs pre-trained on higher-rate ECGs may receive HUCA inputs whose effective spectral support differs from their pre-training distribution, likely contributing to the cross-site performance collapse. Third, our evaluation covers a single rare cardiac condition across two cohorts. The convergence of within-model, between-model, and cross-cohort analyses lends consistency to our conclusions, but their extension to other rare cardiovascular phenotypes remains a question for future work.

## 6. Conclusions

We assessed whether ECG foundation models offer meaningful advantages over supervised baselines for rare cardiac disease detection, using Brugada syndrome as a representative case. Beyond performance, we focused on disentangling the contributions of pre-training versus architecture, quantifying the role of training-set size, and evaluating robustness under cross-site transfer. Across within-model, between-model, and cross-site analyses on nine publicly available FMs, four findings emerged. First, the benefit of pre-training is largely mechanistic: decisive for high-capacity architectures that fail to converge from scratch, but only marginal for compact architectures that already perform well under supervised training. Second, the data-efficiency advantage of pre-training observed on the BrSwiss cohort does not replicate on an independent external cohort. Third, no model and no adaptation strategy achieves reliable cross-site generalization: under zero-shot transfer, performance consistently collapses toward chance level. Finally, architectural suitability emerges as the most consistent determinant of performance across all experimental settings. These results suggest that the primary contribution of current ECG foundation models lies in facilitating optimization and partially mitigating data scarcity, rather than in encoding transferable, clinically meaningful representations. This challenges the prevailing assumption that large-scale pre-training inherently confers generalizable clinical knowledge, particularly for rare diseases. Rather than questioning the utility of FMs altogether, our findings call for a more precise characterization of their limitations and for pre-training strategies explicitly grounded in clinical variability, domain shifts, and disease-specific features. Bridging this gap will be essential for translating foundation models into reliable clinical tools across heterogeneous populations and healthcare settings.

## CRediT authorship contribution statement

**Beatrice Zanchi:** Conceptualization, Methodology, Software, Investigation, Formal analysis, Writing – original draft. **Giuliana Monachino:** Formal analysis, Writing – review & editing. **Alvise Dei Rossi:** Validation, Writing – review & editing. **Luigi Fiorillo:** Validation, Writing – review & editing. **Georgia Sarquella-Brugada:** Investigation, Writing – review & editing. **Giulio Conte:** Resources, Data curation, Validation, Funding acquisition, Writing – review & editing. **Francesca Dalia Faraci:** Supervision, Funding acquisition, Writing – review & editing.

## Funding

B.Z., G.M., G.C. and F.D.F. disclose support for the research of this work from the European Union and the Swiss SERI through the Eurostars funding program (CMIPA project) [grant number E115814].

## References

[1] Awni Y Hannun, Pranav Rajpurkar, Masoumeh Haghpanahi, Geoffrey H Tison, Codie Bourn, Mintu P Turakhia, and Andrew Y Ng. Cardiologist-level arrhythmia detection and classification in ambulatory electrocardiograms using a deep neural network. *Nature medicine*, 25(1):65–69, 2019.

[2] Ramy Elantary and Samar Othman. Artificial intelligence in electrocardiography: From automated arrhythmia detection to predicting hidden cardiovascular disease. *Cureus*, 17(10), 2025.

[3] J Weston Hughes, Jeffrey E Olgin, Robert Avram, Sean A Abreau, Taylor Sittler, Kaahan Radia, Henry Hsia, Tomos Walters, Byron Lee, Joseph E Gonzalez, et al. Performance of a convolutional neural network and explainability technique for 12-lead electrocardiogram interpretation. *JAMA cardiology*, 6(11):1285–1295, 2021.

[4] Hisaki Makimoto, Moritz Höckmann, Tina Lin, David Glöckner, Shqipe Gerguri, Lukas Clasen, Jan Schmidt, Athena Assadi-Schmidt, Alexandru Bejinariu, Patrick Müller, et al. Performance of a convolutional neural network derived from an ecg database in recognizing myocardial infarction. *Scientific reports*, 10(1):8445, 2020.

[5] Brian CS Loh and Patrick HH Then. Deep learning for cardiac computer-aided diagnosis: benefits, issues & solutions. *Mhealth*, 3:45, 2017.

[6] Giuliana Monachino, Beatrice Zanchi, Michael Wand, Giulio Conte, Athina Tzovara, and Francesca Dalia Faraci. Overcoming data scarcity in life-threatening arrhythmia detection through transfer learning. *Communications Medicine*, 5(1):248, 2025.

[7] Allam Jaya Prakash, Abdelkader Nasreddine Belkacem, Ibrahim M Elfadel, Herbert F Jelinek, and Mohamed Atef. Advances in machine and deep learning for ecg beat classification: a systematic review. *Frontiers in Digital Health*, 7:1649923, 2025.

[8] Yu Han, Xiaofeng Liu, Xiang Zhang, and Cheng Ding. Foundation models in electrocardiogram: A review. *arXiv preprint arXiv:2410.19877*, 2024.

[9] Jun Li, Aaron Aguirre, Junior Moura, Che Liu, Lanhai Zhong, Chenxi Sun, Gari Clifford, Brandon Westover, and Shenda Hong. An electrocardiogram foundation model built on over 10 million recordings with external evaluation across multiple domains. *arXiv preprint arXiv:2410.04133*, 2024.

[10] Sehun Kim. Learning general representation of 12-lead electrocardiogram with a joint-embedding predictive architecture. *arXiv preprint arXiv:2410.08559*, 2024.

[11] Yeongyeon Na, Minje Park, Yunwon Tae, and Sunghoon Joo. Guiding masked representation learning to capture spatio-temporal relationship of electrocardiogram. *arXiv preprint arXiv:2402.09450*, 2024.

[12] Che Liu, Zhongwei Wan, Cheng Ouyang, Anand Shah, Wenjia Bai, and Rossella Arcucci. Zero-shot ecg classification with multimodal learning and test-time clinical knowledge enhancement. *arXiv preprint arXiv:2403.06659*, 2024.

[13] Yuanyuan Tian, Zhiyuan Li, Yanrui Jin, Mengxiao Wang, Xiaoyang Wei, Liqun Zhao, Yunqing Liu, Jinlei Liu, and Chengliang Liu. Foundation model of ecg diagnosis: Diagnostics and explanations of any form and rhythm on ecg. *Cell Reports Medicine*, 5(12), 2024.

[14] Edoardo Coppola, Mattia Savardi, Mauro Massussi, Marianna Adamo, Marco Metra, and Alberto Signoroni. Hubert-ecg as a self-supervised foundation model for broad and scalable cardiac applications. *medRxiv*, pages 2024–11, 2024.

[15] Kaden McKeen, Sameer Masood, Augustin Toma, Barry Rubin, and Bo Wang. Ecg-fm: An open electrocardiogram foundation model. *JAMIA open*, 8(5):ooaf122, 2025.

[16] Rishi Bommasani, Drew A Hudson, Ehsan Adeli, Russ Altman, Simran Arora, Sydney von Arx, Michael S Bernstein, Jeannette Bohg, Antoine Bosselut, Emma Brunskill, et al. On the opportunities and risks of foundation models. *arXiv preprint arXiv:2108.07258*, 2021.

[17] M A Al-Masud, Juan Lopez Alcaraz, and Nils Strodthoff. Benchmarking ECG FMs: A reality check across clinical tasks. In *The Fourteenth International Conference on Learning Representations*, 2026.

[18] Zhijiang Wan, Qianhao Yu, Jia Mao, Wenfeng Duan, and Cheng Ding. Openecg: Benchmarking ecg foundation models with public 1.2 million records. *arXiv preprint arXiv:2503.00711*, 2025.

[19] Yuhao Xu, Jiaying Lu, Sirui Ding, Defu Cao, Xiao Hu, and Carl Yang. An electrocardiogram multi-task benchmark with comprehensive evaluations and insightful findings. *arXiv preprint arXiv:2512.08954*, 2025.

[20] Riccardo Lunelli, Angus Nicolson, Samuel Martin Pröll, Sebastian Johannes Reinstadler, Axel Bauer, and Clemens Dlaska. Benchecg and xecg: a benchmark and baseline for ecg foundation models. *arXiv preprint arXiv:2509.10151*, 2025.

[21] Stephanie M Hu, Joshua P Barrios, and Geoffrey H Tison. A deep foundation model for electrocardiogram interpretation: enabling rare disease detection through transfer learning. *European Heart Journal-Digital Health*, 6(4):619–623, 2025.

[22] Robert Ronan, Constantine Tarabanis, Larry Chinitz, and Lior Jankelson. Self-supervised vicreg pre-training for brugada ecg detection. *Scientific Reports*, 15(1):9396, 2025.

[23] Satoshi Kawada, Hiroshi Morita, Charles Antzelevitch, Yoshimasa Morimoto, Koji Nakagawa, Atsuyuki Watanabe, Nobuhiro Nishii, Kazufumi Nakamura, and Hiroshi Ito. Shanghai score system for diagnosis of brugada syndrome: validation of the score system and system and reclassification of the patients. *Clinical Electrophysiology*, 4(6):724–730, 2018.

[24] Nahuel Costa Cortez and Daniel Garcia Iglesias. Brugada-HUCA: 12-Lead ECG Recordings for the Study of Brugada Syndrome. *PhysioNet*, February 2026. Version 1.0.0.

[25] Nils Strodthoff, Juan Miguel Lopez Alcaraz, and Wilhelm Haverkamp. Prospects for artificial intelligence-enhanced electrocardiogram as a unified screening tool for cardiac and non-cardiac conditions: an explorative study in emergency care. *European Heart Journal-Digital Health*, 5(4):454–460, 2024.

[26] Kaiming He, Xiangyu Zhang, Shaoqing Ren, and Jian Sun. Deep residual learning for image recognition. In *Proceedings of the IEEE conference on computer vision and pattern recognition*, pages 770–778, 2016.

[27] Wei-Ning Hsu, Benjamin Bolte, Yao-Hung Hubert Tsai, Kushal Lakhotia, Ruslan Salakhutdinov, and Abdelrahman Mohamed. Hubert: Self-supervised speech representation learning by masked prediction of hidden units. *IEEE/ACM transactions on audio, speech, and language processing*, 29:3451–3460, 2021.

[28] Ashish Vaswani, Noam Shazeer, Niki Parmar, Jakob Uszkoreit, Llion Jones, Aidan N Gomez, Łukasz Kaiser, and Illia Polosukhin. Attention is all you need. *Advances in neural information processing systems*, 30, 2017.

[29] Alexei Baevski, Yuhao Zhou, Abdelrahman Mohamed, and Michael Auli. wav2vec 2.0: A framework for self-supervised learning of speech representations. *Advances in neural information processing systems*, 33:12449–12460, 2020.

[30] Dani Kiyasseh, Tingting Zhu, and David A Clifton. Clocs: Contrastive learning of cardiac signals across space, time, and patients. In *International Conference on Machine Learning*, pages 5606–5615. PMLR, 2021.

[31] Alexey Dosovitskiy. An image is worth 16x16 words: Transformers for image recognition at scale. *arXiv preprint arXiv:2010.11929*, 2020.

[32] Yann LeCun. A path towards autonomous machine intelligence version 0.9. 2, 2022-06-27. *Open Review*, 62(1):1–62, 2022.

[33] Kaiming He, Xinlei Chen, Saining Xie, Yanghao Li, Piotr Dollár, and Ross Girshick. Masked autoencoders are scalable vision learners. In *Proceedings of the IEEE/CVF conference on computer vision and pattern recognition*, pages 16000–16009, 2022.

[34] Albert Gu, Karan Goel, and Christopher Ré. Efficiently modeling long sequences with structured state spaces. *arXiv preprint arXiv:2111.00396*, 2021.

[35] Aaron van den Oord, Yazhe Li, and Oriol Vinyals. Representation learning with contrastive predictive coding. *arXiv preprint arXiv:1807.03748*, 2018.

[36] Shenda Hong, Yanbo Xu, Alind Khare, Satria Priambada, Kevin Maher, Alaa Aljiffry, Jimeng Sun, and Alexey Tumanov. Holmes: health online model ensemble serving for deep learning models in intensive care units. In *Proceedings of the 26th ACM SIGKDD International Conference on Knowledge Discovery & Data Mining*, pages 1614–1624, 2020.

[37] Ilija Radosavovic, Raj Prateek Kosaraju, Ross Girshick, Kaiming He, and Piotr Dollár. Designing network design spaces. In *Proceedings of the IEEE/CVF conference on computer vision and pattern recognition*, pages 10428–10436, 2020.

[38] Yunrui Zhao, Qianqian Xu, Yangbangyan Jiang, Peisong Wen, and Qingming Huang. Dist-pu: Positive-unlabeled learning from a label distribution perspective. In *Proceedings of the IEEE/CVF Conference on Computer Vision and Pattern Recognition*, pages 14461–14470, 2022.

[39] Nils Strodthoff, Patrick Wagner, Tobias Schaeffter, and Wojciech Samek. Deep learning for ecg analysis: Benchmarks and insights from ptb-xl. *IEEE journal of biomedical and health informatics*, 25(5):1519–1528, 2020.

[40] Hangbo Bao, Li Dong, Songhao Piao, and Furu Wei. Beit: Bert pre-training of image transformers. *arXiv preprint arXiv:2106.08254*, 2021.

[41] Diederik Kinga, Jimmy Ba Adam, et al. A method for stochastic optimization. In *International conference on learning representations (ICLR)*, volume 5. California;, 2015.

[42] Hugo Touvron, Matthieu Cord, Matthijs Douze, Francisco Massa, Alexandre Sablayrolles, and Hervé Jégou. Training data-efficient image transformers & distillation through attention. In *International conference on machine learning*, pages 10347–10357. PMLR, 2021.

[43] Stéphane d'Ascoli, Hugo Touvron, Matthew L Leavitt, Ari S Morcos, Giulio Biroli, and Levent Sagun. Convit: Improving vision transformers with soft convolutional inductive biases. In *International conference on machine learning*, pages 2286–2296. PMLR, 2021.

[44] Jared Kaplan, Sam McCandlish, Tom Henighan, Tom B Brown, Benjamin Chess, Rewon Child, Scott Gray, Alec Radford, Jeffrey Wu, and Dario Amodei. Scaling laws for neural language models. *arXiv preprint arXiv:2001.08361*, 2020.

[45] Jordan Hoffmann, Sebastian Borgeaud, Arthur Mensch, Elena Buchatskaya, Trevor Cai, Eliza Rutherford, Diego de Las Casas, Lisa Anne Hendricks, Johannes Welbl, Aidan Clark, et al. Training compute-optimal large language models. *arXiv preprint arXiv:2203.15556*, 2022.

[46] Iulian Emil Tampu, Anders Eklund, and Neda Haj-Hosseini. Inflation of test accuracy due to data leakage in deep learning-based classification of oct images. *Scientific Data*, 9(1):580, 2022.

[47] Mukund Sundararajan, Ankur Taly, and Qiqi Yan. Axiomatic attribution for deep networks. In *International conference on machine learning*, pages 3319–3328. PMLR, 2017.

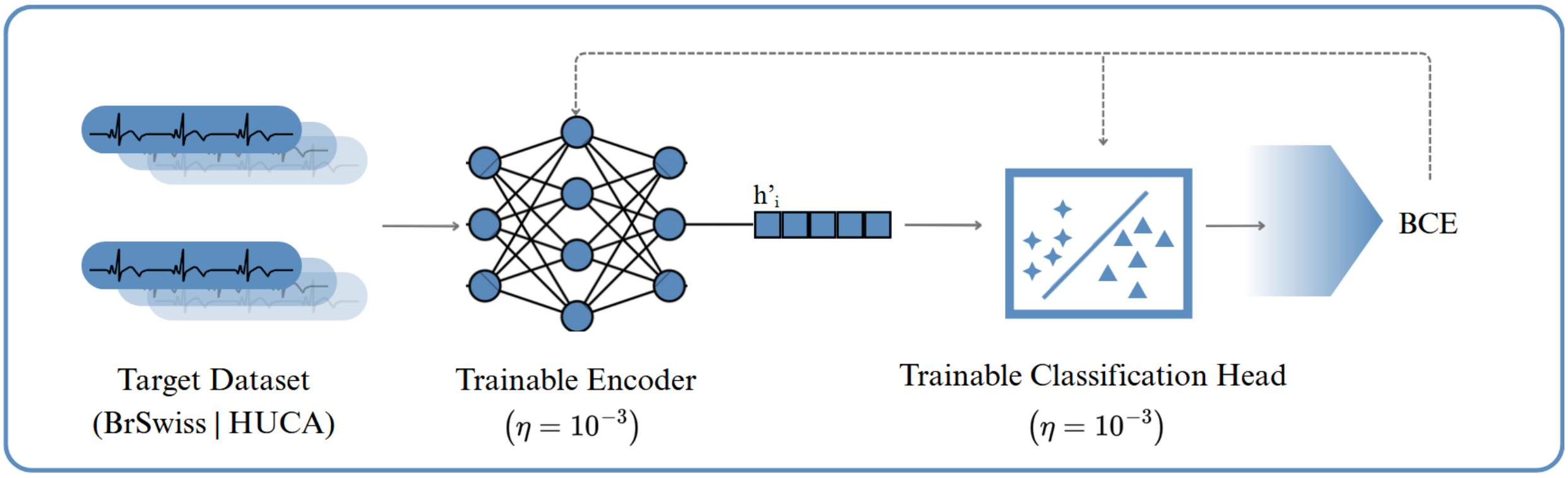

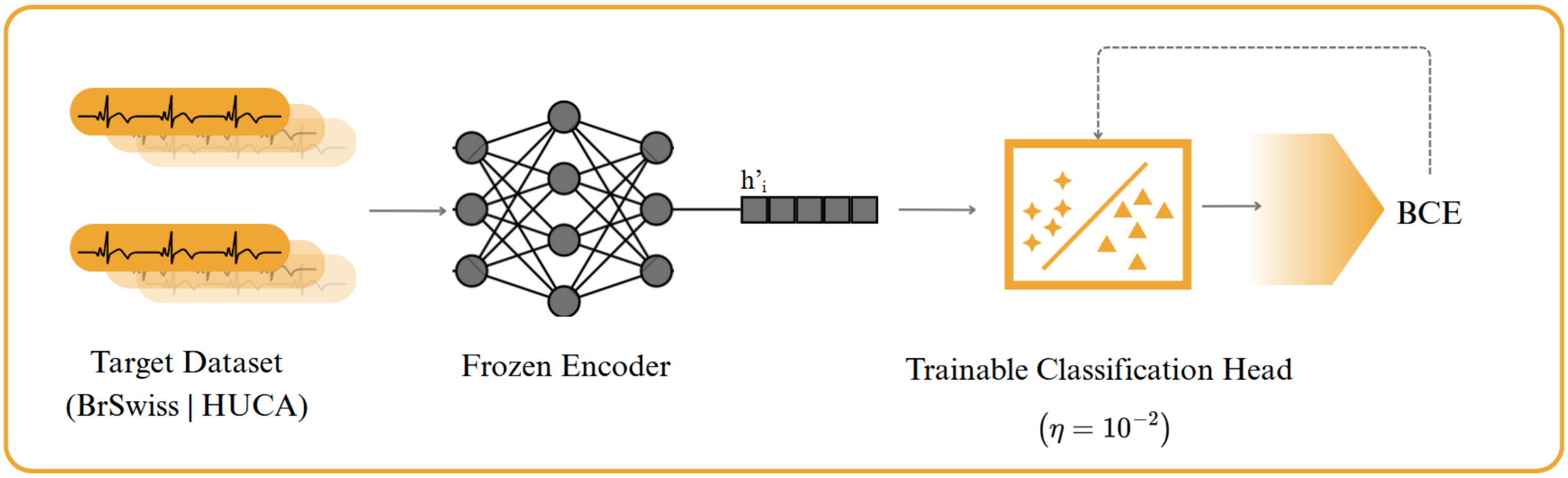

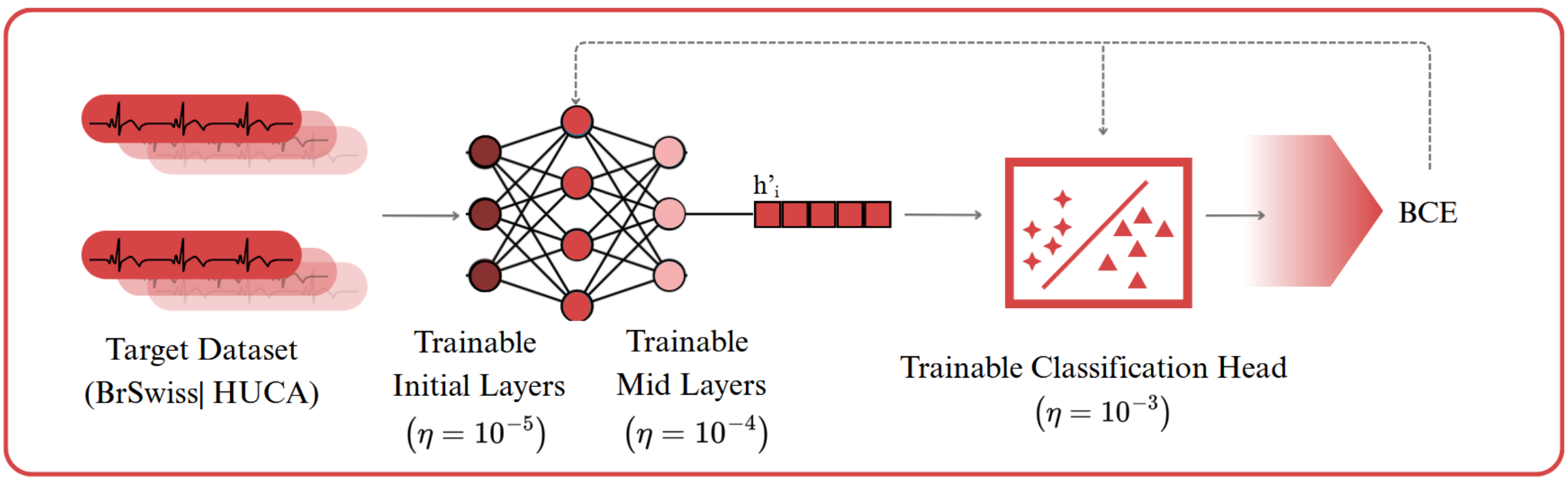

**Figure 1:** Schematic overview of the experimental settings in our study. In the Train from Scratch approach (a), the model is initialized with random weights. In the Linear Evaluation approach (b), the pre-trained encoder weights are kept frozen to act as a static feature extractor, while only the classification head is trained. The Full Fine-Tuning approach (c) involves updating the entire pre-trained network. All strategies utilize the Adam optimizer to minimize Binary Cross-Entropy loss.

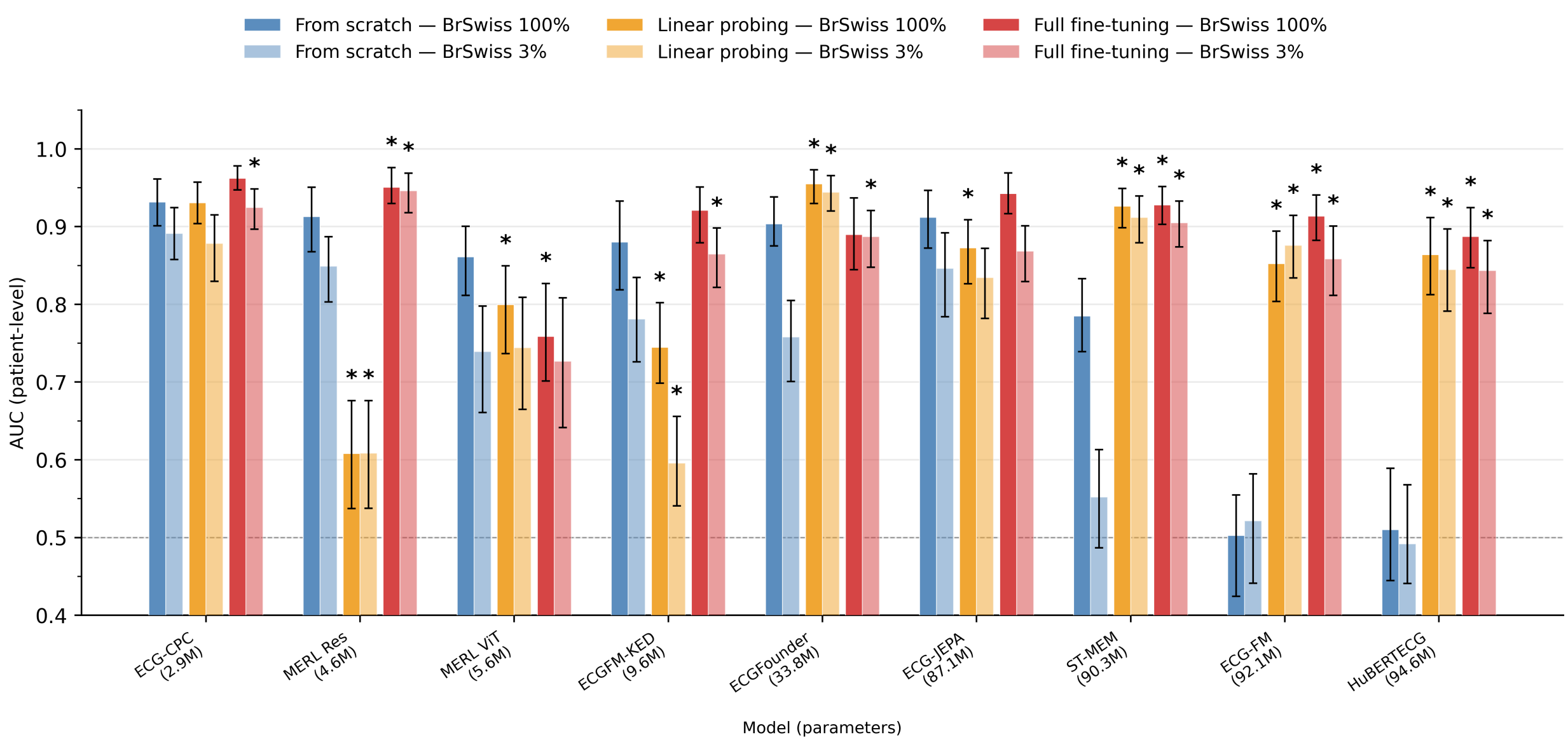

**Figure 2:** Patient-level AUC for the nine ECG foundation models tested on the same BrSwiss test set. Each model group contains six bars representing the three evaluation strategies (blue = From scratch training; orange = Linear probing; and red = Full fine-tuning) trained on both the full BrSwiss-100% training set (full opacity) and a reduced 3% subset (reduced opacity). Models are ordered by parameter count. Error bars denote 95% bootstrap confidence intervals. Asterisks (*) indicate a statistically significant difference between the adaptation strategy (Linear probing or Full fine-tuning) and the corresponding from-scratch baseline of the same architecture (within-model comparison; DeLong's test, FDR-adjusted $p < 0.05$).

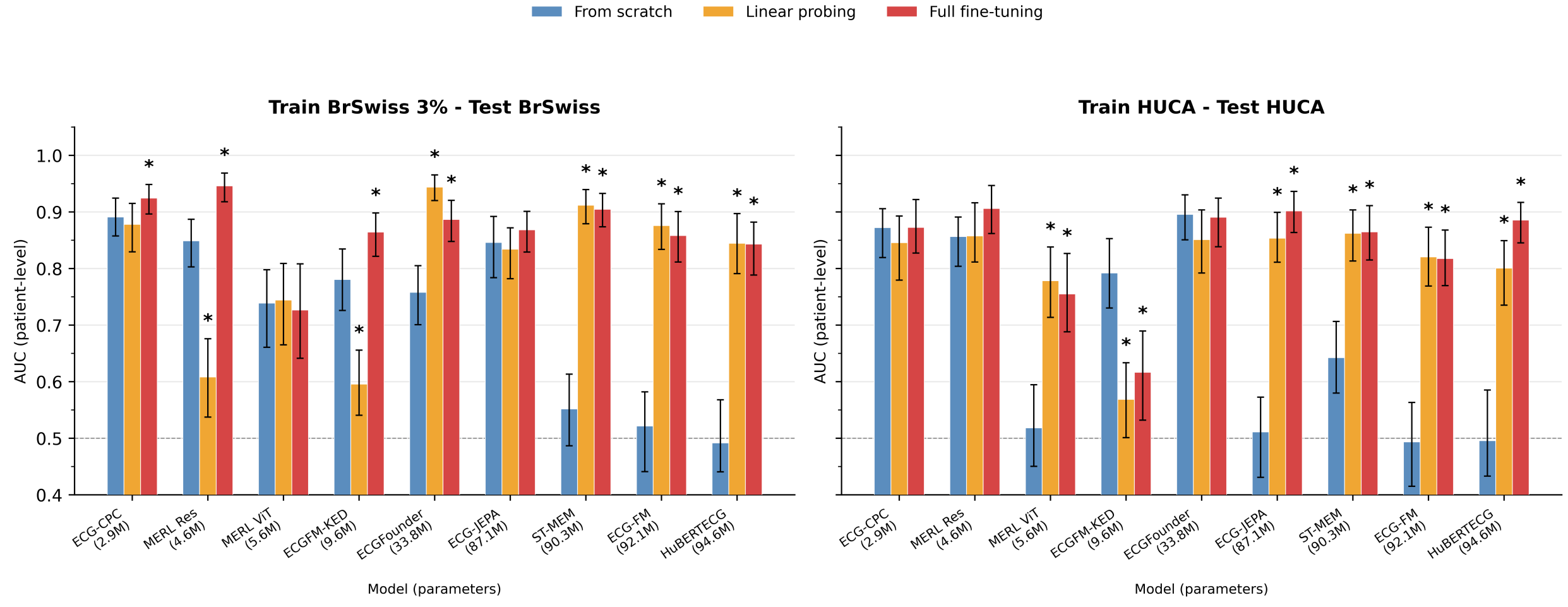

**Figure 3:** Patient-level AUC for the nine ECG foundation models under the three adaptation strategies (blue = From scratch training; orange = Linear probing; and red = Full fine-tuning). The left panel shows models trained and tested on the reduced BrSwiss training set (BrSwiss-3%); the right panel shows models trained and tested on the HUCA dataset. Models are ordered by parameter count. Error bars denote 95% bootstrap confidence intervals. Asterisks (*) indicate a statistically significant difference between the adaptation strategy (Linear probing or Full fine-tuning) and the corresponding from-scratch baseline of the same architecture (within-model comparison; DeLong's test, FDR-adjusted $p < 0.05$).

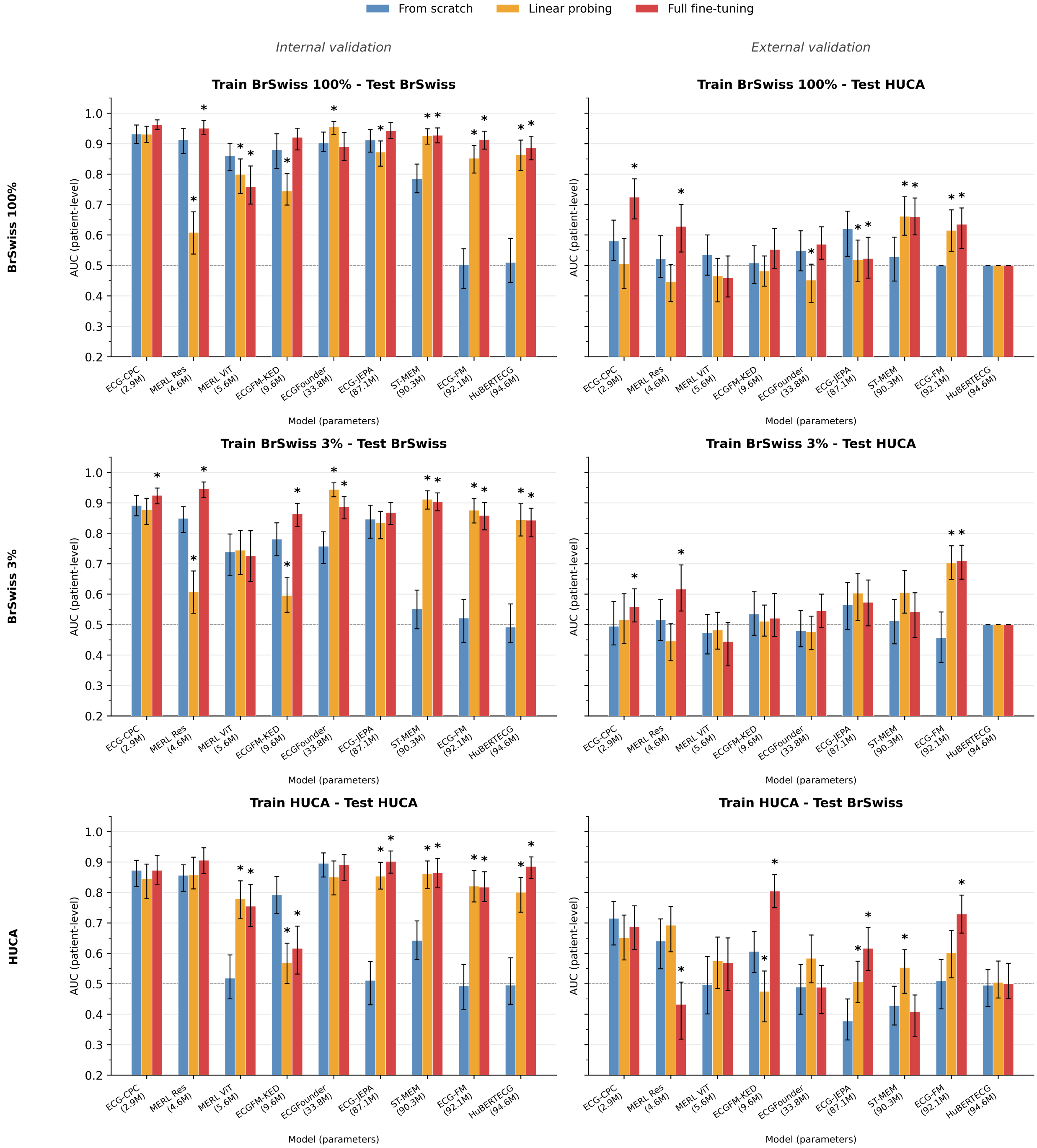

**Figure 4:** Patient-level AUC for the nine ECG foundation models across all six experimental configurations, under the three adaptation strategies (blue = From scratch training; orange = Linear probing; and red = Full fine-tuning). Rows correspond to the training source: the full BrSwiss training set (BrSwiss-100%, top), the reduced BrSwiss training set (BrSwiss-3 %), and the HUCA training set (bottom). The left column reports BrSwiss validation (test set drawn from the same cohort used for training) and the right column reports external validation (zero-shot cross-site transfer to the other cohort, without additional fine-tuning). Models are ordered by parameter count. Error bars denote 95% bootstrap confidence intervals. Asterisks (*) indicate a statistically significant difference between the adaptation strategy (Linear probing or Full fine-tuning) and the corresponding from-scratch baseline of the same architecture (within-model comparison; DeLong's test, FDR-adjusted $p < 0.05$).

## A. Detailed Data Collection and Preprocessing

ECG recordings from the BrSwiss dataset are obtained using high-resolution ECG machines (CARDIOVIT CS-200 Excellence; Schiller) with a sampling frequency of 1000 Hz. Each recording consists of 12 leads collected under resting conditions, with durations ranging from 15 seconds to 5 minutes. The preprocessing pipeline consists of two main steps applied uniformly across all models and both datasets (BrSwiss and HUCA). First, noise reduction is performed using discrete wavelet transform with Daubechies 4 wavelet and soft thresholding (threshold = 50 $\mu$V). Second, each recording is resampled to the target sampling frequency required by the native input specifications of each FM package (ranging from 100 Hz to 500 Hz) using time-domain resampling. This resampling step is performed after denoising to preserve signal integrity at the highest available resolution before downsampling.

Both the BrSwiss and HUCA datasets are partitioned using a stratified 3-fold cross-validation scheme, preserving class balance across splits (29% and 20.9% for BrS subjects, respectively). In each iteration, one fold is held out for testing, while the remaining two folds are further divided into 80% for training and 20% for validation. To meet the specific input requirements of different FMs, the full-length ECG recordings are windowed into segments of fixed duration. This segmentation process results in multiple segments per patient. To strictly prevent data leakage, all segments derived from a single patient are constrained to the same data split (training, validation, or testing), ensuring that the model never encounters segments from a test patient during the training phase [46]. Concerning ablation studies, we further reduce the BrSwiss training and validation sets to 3% of the original length while keeping the test sets fixed.

## B. Detailed layer-wise learning rate decay configurations

To optimize the fine-tuning of diverse FMs, we employ a LLRD strategy. As shown in Table 5, we partition each architecture into three distinct regions (Initial, Medium, and Top) with learning rates $\eta_{low}$, $\eta_{mid}$, and $\eta_{high}$ respectively. In our experiments, $\eta_{high} = 10^{-3}$, $\eta_{mid} = 10^{-4}$, and $\eta_{low} = 10^{-5}$.

## C. Detailed Evaluation Protocols and Statistical Analysis

All models are evaluated on identical train-validation-test splits for both the BrSwiss and HUCA datasets. The downstream task is formulated as a binary classification problem, where the objective is to distinguish between control group subjects and patients diagnosed with BrS. Performances are reported at the patient-level, where metrics are computed by aggregating segment-level prediction scores within each patient via mean probability pooling, yielding a single score per patient. For threshold-dependent metrics (accuracy, sensitivity, specificity, and F1-score), a fixed decision threshold of 0.5 is applied uniformly across all models and tasks. The area under the receiver operating characteristic curve (AUC) is used as the primary performance metric for model comparison. Complete evaluation metrics results and saliency maps analysis are reported in the Appendix. To quantify uncertainty in performance estimates and to formally compare models, patient-level predictions from the three cross-validation folds were concatenated into a single pooled out-of-fold prediction vector per model, on which a single AUC was computed. This ensures that every patient contributes exactly once to the evaluation, providing identical paired predictions across models. 95% confidence intervals for all metrics were obtained via non-parametric bootstrap resampling of this pooled vector at the patient level (2000 iterations). Statistical significance of AUC differences was assessed via DeLong's test for paired AUC, applied to the pooled out-of-fold predictions. To support both the within-model and between-model analyses, comparisons were structured as follows: for each adaptation strategy (Linear Probing, Full Fine-Tuning), every foundation model was tested against each of the nine supervised from-scratch baselines, yielding 81 paired comparisons per [strategy, experimental configuration] block. The Benjamini-Hochberg procedure was applied to control the False Discovery Rate independently within each (strategy, experimental configuration) block of 81 p-values. Differences were considered statistically significant at an FDR-adjusted p-value$< 0.05$.

## D. Detailed results

This appendix reports the complete patient-level results underlying the figures presented in the main text. The first group of tables provides the per-model AUC values. Tables 6 and 7 correspond to the data-efficiency analysis, reporting respectively the reduced BrSwiss training set (BrSwiss-3%) and the HUCA cohort, each evaluated within-site under the three adaptation strategies. Tables 8, 9, and 10 report the three zero-shot cross-site configurations: training on BrSwiss-100% and testing on HUCA, training on BrSwiss-3% and testing on HUCA, and training on HUCA and testing on BrSwiss, respectively.

The second group of tables (Tables 11–16) reports the secondary classification metrics (accuracy, sensitivity, specificity, precision, and F1) for all six experimental configurations: the three within-site settings (BrSwiss-100%, BrSwiss-3%, and HUCA) and the three cross-site settings. In these tables, results are grouped by adaptation strategy, and the best value per metric within each strategy is shown in bold, excluding models with degenerate (single-class) predictions. All threshold-dependent metrics are computed at a fixed decision threshold of 0.5.

Across all tables, values are reported as mean with 95% bootstrap confidence intervals, aggregated at the patient level across the three cross-validation folds. Where shown, asterisks denote within-model statistical significance relative to the from-scratch baseline of the same architecture (DeLong's test, FDR-adjusted $p < 0.05$).

## E. Comparative saliency analysis across models

To provide a qualitative view of the temporal regions each model attends to when predicting Brugada syndrome, we computed Integrated Gradients (IG) [47] attributions on a small selection of test ECGs. For each sample, IG estimates the contribution of every input sample to the predicted class score by integrating the gradient of the score with respect to the input along a straight-line path from a zero baseline to the actual signal. Attributions were computed with 100 interpolation steps and projected onto the 12 leads of the original ECG. Figures 5, 6, 7 show representative examples for each model under the three adaptation strategies (FS, LP, FT).

The saliency maps of from-scratch models tend to appear visually more saturated than those of their linear-probed or fine-tuned counterparts. This difference does not imply that from-scratch models are more confident, more focused, or more clinically interpretable. It largely reflects the fact that randomly initialised classification heads, trained directly on a small labelled set, tend to produce logits with larger absolute values than heads operating on pre-trained representations, and the magnitude of IG attributions scales linearly with the magnitude of the logit being attributed. Each map in our figures is normalised by its own maximum absolute value for display, so the colour intensity within a single panel encodes the relative spatial distribution of the attribution, not its absolute scale. Comparisons of visual intensity across panels should therefore be interpreted with caution.

**Table 5**
Detailed LLRD partitioning per model architecture with specific learning rate.

| Model | Initial ($\eta = 10^{-5}$) | Medium ($\eta = 10^{-4}$) | Top ($\eta = 10^{-3}$) |
|---|---|---|---|
| ECG-CPC | `encoder_module` | `predictor_module` | Head |
| ECG-FM | `feat_ext`, Blocks 0–5 | Blocks 6–11, `final_proj` | Head |
| ECGFounder | `first_conv`, Stages 0–2 | Stages 3–6 | Head |
| ECG-JEPA | `W_P`, Blocks 0–2 | Blocks 3–11, `norm` | Head |
| ECGFM-KED | Stages 0–2 | Stages 4–7 | Head |
| HuBERT-ECG | `feat_ext`, Blocks $0 \ldots (\frac{N}{2} - 1)$ | Blocks $\frac{N}{2} \ldots (N - 1)$ | Head |
| MERL ViT | `patch_embed`, Blocks $0 \ldots (\frac{N}{2} - 1)$ | Blocks $\frac{N}{2} \ldots (N - 1)$, `norm` | Head |
| MERL ResNet | `conv1`, Layers 1–2 | Layers 3–4 | Head |
| ST-MEM | `pos_embed`, Blocks 0–2 | Blocks 3–11, `norm` | Head |

**Table 6**
Patient-level AUC (mean [95% CI]) on the BrSwissdataset (3% training windows) across three experimental settings.

| Model | FS | LP | FT |
|---|---|---|---|
| HuBERTECG (94.6 M) | 0.492 [0.441–0.568] | 0.845*[0.791–0.897] | 0.843*[0.788–0.882] |
| ECG-FM (92.1 M) | 0.522 [0.441–0.582] | 0.876*[0.834–0.915] | 0.858*[0.811–0.901] |
| ST-MEM (90.3 M) | 0.552 [0.487–0.613] | 0.912*[0.879–0.940] | 0.905*[0.874–0.933] |
| ECG-JEPA (87.1 M) | 0.846 [0.784–0.892] | 0.834 [0.782–0.872] | 0.868 [0.829–0.901] |
| ECGFounder (33.8 M) | 0.758 [0.701–0.805] | 0.944*[0.920–0.966] | 0.887*[0.848–0.921] |
| ECGFM-KED (9.6 M) | 0.781 [0.726–0.835] | 0.596*[0.541–0.656] | 0.865*[0.822–0.898] |
| MERL-ViT (5.6 M) | 0.739 [0.661–0.798] | 0.744 [0.665–0.809] | 0.727 [0.641–0.808] |
| MERL-ResNet (4.6 M) | 0.849 [0.803–0.887] | 0.608*[0.538–0.676] | 0.946*[0.918–0.969] |
| ECG-CPC (2.9 M) | 0.891 [0.858–0.924] | 0.878 [0.830–0.915] | 0.924*[0.897–0.949] |

**Training strategies**: FS = from scratch; LP = Linear Probing; FT = Full Fine-Tuning.
*****: Statistically significant difference compared to the supervised baseline trained from scratch (DeLong's test, FDR-adjusted $p < 0.05$).

**Table 7**
Patient-level AUC (mean [95% CI]) on the HUCA dataset across three experimental settings.

| Model | FS | LP | FT |
|---|---|---|---|
| HuBERTECG (94.6 M) | 0.496 [0.433–0.585] | 0.801*[0.735–0.849] | 0.886*[0.845–0.917] |
| ECG-FM (92.1 M) | 0.493 [0.415–0.563] | 0.821*[0.769–0.873] | 0.818*[0.770–0.868] |
| ST-MEM (90.3 M) | 0.643 [0.580–0.707] | 0.862*[0.813–0.904] | 0.865*[0.815–0.911] |
| ECG-JEPA (87.1 M) | 0.511 [0.431–0.573] | 0.854*[0.811–0.899] | 0.902*[0.864–0.937] |
| ECGFounder (33.8 M) | 0.896 [0.851–0.930] | 0.851 [0.792–0.904] | 0.891 [0.839–0.924] |
| ECGFM-KED (9.6 M) | 0.792 [0.730–0.853] | 0.569*[0.501–0.634] | 0.617*[0.532–0.690] |
| MERL-ViT (5.6 M) | 0.518 [0.451–0.595] | 0.779*[0.714–0.838] | 0.755*[0.688–0.827] |
| MERL-ResNet (4.6 M) | 0.857 [0.804–0.891] | 0.858 [0.812–0.916] | 0.906 [0.862–0.947] |
| ECG-CPC (2.9 M) | 0.873 [0.820–0.906] | 0.846 [0.779–0.893] | 0.873 [0.827–0.922] |

**Training strategies**: FS = from scratch; LP = Linear Probing; FT = Full Fine-Tuning.
*****: Statistically significant difference compared to the supervised baseline trained from scratch (DeLong's test, FDR-adjusted $p < 0.05$).

**Table 8**
Cross-site performance - train BrSwiss (100% training windows), test HUCA. Patient-level AUC (mean [95% CI]).

| Model | FS | LP | FT |
|---|---|---|---|
| HuBERTECG (94.6 M) | 0.500 [0.500–0.500] | 0.500 [0.500–0.500] | 0.500 [0.500–0.500] |
| ECG-FM (92.1 M) | 0.500 [0.500–0.500] | 0.615*[0.547–0.682] | 0.635*[0.555–0.689] |
| ST-MEM (90.3 M) | 0.528 [0.449–0.593] | 0.661*[0.599–0.726] | 0.660*[0.601–0.722] |
| ECG-JEPA (87.1 M) | 0.620 [0.530–0.679] | 0.519*[0.447–0.584] | 0.523*[0.458–0.593] |
| ECGFounder (33.8 M) | 0.549 [0.483–0.614] | 0.452*[0.378–0.504] | 0.570 [0.520–0.627] |
| ECGFM-KED (9.6 M) | 0.508 [0.441–0.565] | 0.482 [0.432–0.531] | 0.552 [0.489–0.622] |
| MERL-ViT (5.6 M) | 0.536 [0.469–0.600] | 0.466 [0.380–0.523] | 0.459 [0.396–0.531] |
| MERL-ResNet (4.6 M) | 0.522 [0.461–0.597] | 0.446 [0.381–0.503] | 0.629*[0.544–0.701] |
| ECG-CPC (2.9 M) | 0.580 [0.516–0.649] | 0.505 [0.424–0.589] | 0.724*[0.653–0.784] |

**Training strategies**: FS = from scratch; LP = Linear Probing; FT = Full Fine-Tuning.
*****: Statistically significant difference compared to the supervised baseline trained from scratch (DeLong's test, FDR-adjusted $p < 0.05$).

**Table 9**
Cross-site performance - train BrSwiss (3% training windows), test HUCA. Patient-level AUC (mean [95% CI]).

| Model | FS | LP | FT |
|---|---|---|---|
| HuBERTECG (94.6 M) | 0.500 [0.500–0.500] | 0.500 [0.500–0.500] | 0.500 [0.500–0.500] |
| ECG-FM (92.1 M) | 0.456 [0.376–0.542] | 0.703*[0.648–0.759] | 0.710*[0.649–0.761] |
| ST-MEM (90.3 M) | 0.513 [0.437–0.583] | 0.605 [0.538–0.678] | 0.542 [0.457–0.605] |
| ECG-JEPA (87.1 M) | 0.565 [0.484–0.638] | 0.604 [0.514–0.667] | 0.573 [0.496–0.647] |
| ECGFounder (33.8 M) | 0.479 [0.428–0.546] | 0.476 [0.418–0.528] | 0.546 [0.489–0.600] |
| ECGFM-KED (9.6 M) | 0.535 [0.465–0.608] | 0.511 [0.463–0.564] | 0.521 [0.461–0.602] |
| MERL-ViT (5.6 M) | 0.473 [0.404–0.533] | 0.483 [0.420–0.541] | 0.445 [0.365–0.507] |
| MERL-ResNet (4.6 M) | 0.516 [0.449–0.582] | 0.446 [0.381–0.503] | 0.616*[0.545–0.696] |
| ECG-CPC (2.9 M) | 0.494 [0.433–0.576] | 0.516 [0.438–0.602] | 0.558*[0.509–0.618] |

**Training strategies**: FS = from scratch; LP = Linear Probing; FT = Full Fine-Tuning.
*****: Statistically significant difference compared to the supervised baseline trained from scratch (DeLong's test, FDR-adjusted $p < 0.05$).

**Table 10**
Cross-site performance - train HUCA, test BrSwiss. Patient-level AUC (mean [95% CI]).

| Model | FS | LP | FT |
|---|---|---|---|
| HuBERTECG (94.6 M) | 0.495 [0.425–0.547] | 0.505 [0.453–0.575] | 0.501 [0.451–0.568] |
| ECG-FM (92.1 M) | 0.509 [0.418–0.580] | 0.602 [0.520–0.676] | 0.729*[0.667–0.791] |
| ST-MEM (90.3 M) | 0.428 [0.365–0.492] | 0.554*[0.469–0.612] | 0.409 [0.328–0.463] |
| ECG-JEPA (87.1 M) | 0.378 [0.315–0.450] | 0.508*[0.438–0.575] | 0.617*[0.544–0.684] |
| ECGFounder (33.8 M) | 0.489 [0.400–0.564] | 0.584 [0.503–0.660] | 0.489 [0.402–0.561] |
| ECGFM-KED (9.6 M) | 0.606 [0.537–0.672] | 0.475*[0.375–0.542] | 0.804*[0.750–0.859] |
| MERL-ViT (5.6 M) | 0.497 [0.401–0.589] | 0.576 [0.484–0.653] | 0.569 [0.478–0.651] |
| MERL-ResNet (4.6 M) | 0.641 [0.549–0.713] | 0.692 [0.605–0.754] | 0.432*[0.318–0.506] |
| ECG-CPC (2.9 M) | 0.715 [0.628–0.770] | 0.652 [0.578–0.726] | 0.688 [0.612–0.756] |

**Training strategies**: FS = from scratch; LP = Linear Probing; FT = Full Fine-Tuning.
*****: Statistically significant difference compared to the supervised baseline trained from scratch (DeLong's test, FDR-adjusted $p < 0.05$).

Table 11: Patient-level secondary metrics on the BrSwiss-100% dataset (within-site).

| | Model | Accuracy | Sensitivity | Specificity | Precision | F1 |
|---|---|---|---|---|---|---|
| FS | HuBERTECG (94.6 M) | 0.706 [0.661–0.769] | 0.000 [0.000–0.000] | 1.000 [1.000–1.000] | 0.000 [0.000–0.000] | 0.000 [0.000–0.000] |
| | ECG-FM (92.1 M) | 0.706 [0.661–0.769] | 0.000 [0.000–0.000] | 1.000 [1.000–1.000] | 0.000 [0.000–0.000] | 0.000 [0.000–0.000] |
| | ST-MEM (90.3 M) | 0.702 [0.657–0.767] | 0.000 [0.000–0.000] | 0.995 [0.983–1.000] | 0.000 [0.000–0.000] | 0.000 [0.000–0.000] |
| | ECG-JEPA (87.1 M) | 0.866 [0.825–0.904] | 0.663 [0.576–0.768] | 0.952 [0.928–0.976] | 0.851 [0.774–0.926] | 0.745 [0.679–0.815] |
| | ECGFounder (33.8 M) | 0.819 [0.781–0.868] | **0.791 [0.718–0.902]** | 0.830 [0.782–0.884] | 0.660 [0.570–0.759] | 0.720 [0.659–0.790] |
| | ECGFM-KED (9.6 M) | 0.836 [0.793–0.880] | 0.570 [0.459–0.678] | 0.947 [0.914–0.973] | 0.817 [0.731–0.905] | 0.671 [0.588–0.755] |
| | MERL-ViT (5.6 M) | 0.819 [0.777–0.865] | 0.454 [0.337–0.547] | **0.971 [0.952–0.990]** | **0.867 [0.767–0.957]** | 0.595 [0.472–0.676] |
| | MERL-ResNet (4.6 M) | 0.866 [0.829–0.901] | 0.744 [0.639–0.829] | 0.917 [0.887–0.954] | 0.790 [0.707–0.877] | 0.766 [0.705–0.827] |
| | ECG-CPC (2.9 M) | **0.890 [0.866–0.918]** | 0.767 [0.689–0.866] | 0.942 [0.916–0.971] | 0.846 [0.769–0.916] | **0.805 [0.748–0.860]** |
| LP | HuBERTECG (94.6 M) | 0.822 [0.781–0.863] | 0.581 [0.506–0.697] | 0.922 [0.883–0.954] | 0.758 [0.641–0.847] | 0.658 [0.575–0.737] |
| | ECG-FM (92.1 M) | 0.808 [0.758–0.850] | 0.744 [0.646–0.828] | 0.835 [0.780–0.879] | 0.653 [0.529–0.744] | 0.696 [0.598–0.768] |
| | ST-MEM (90.3 M) | 0.860 [0.822–0.897] | 0.791 [0.714–0.869] | 0.888 [0.845–0.927] | 0.747 [0.645–0.830] | 0.768 [0.688–0.833] |
| | ECG-JEPA (87.1 M) | 0.812 [0.770–0.855] | 0.616 [0.511–0.709] | 0.893 [0.849–0.932] | 0.707 [0.588–0.795] | 0.658 [0.571–0.731] |
| | ECGFounder (33.8 M) | **0.907 [0.875–0.938]** | 0.837 [0.756–0.909] | 0.937 [0.902–0.966] | 0.847 [0.765–0.916] | **0.842 [0.784–0.886]** |
| | ECGFM-KED (9.6 M) | 0.603 [0.548–0.668] | **0.884 [0.817–0.938]** | 0.485 [0.432–0.564] | 0.418 [0.331–0.473] | 0.567 [0.480–0.629] |
| | MERL-ViT (5.6 M) | 0.692 [0.639–0.737] | 0.721 [0.620–0.813] | 0.680 [0.615–0.748] | 0.484 [0.374–0.574] | 0.579 [0.483–0.637] |
| | MERL-ResNet (4.6 M) | 0.671 [0.625–0.723] | 0.023 [0.000–0.058] | 0.942 [0.908–0.966] | 0.143 [0.000–0.265] | 0.040 [0.000–0.093] |
| | ECG-CPC (2.9 M) | 0.870 [0.825–0.906] | 0.674 [0.566–0.771] | **0.952 [0.923–0.972]** | **0.853 [0.754–0.916]** | 0.753 [0.659–0.826] |
| FT | HuBERTECG (94.6 M) | 0.849 [0.813–0.884] | 0.814 [0.741–0.900] | 0.864 [0.833–0.902] | 0.714 [0.633–0.781] | 0.761 [0.703–0.816] |
| | ECG-FM (92.1 M) | 0.846 [0.813–0.884] | 0.779 [0.703–0.857] | 0.874 [0.826–0.913] | 0.720 [0.609–0.807] | 0.749 [0.667–0.815] |
| | ST-MEM (90.3 M) | 0.829 [0.784–0.875] | 0.802 [0.729–0.882] | 0.840 [0.796–0.895] | 0.676 [0.557–0.775] | 0.734 [0.634–0.802] |
| | ECG-JEPA (87.1 M) | 0.904 [0.872–0.942] | 0.721 [0.655–0.835] | **0.981 [0.956–0.995]** | **0.939 [0.853–0.986]** | 0.816 [0.749–0.885] |
| | ECGFounder (33.8 M) | 0.839 [0.791–0.880] | 0.756 [0.667–0.847] | 0.874 [0.828–0.921] | 0.714 [0.635–0.798] | 0.735 [0.674–0.808] |
| | ECGFM-KED (9.6 M) | 0.873 [0.832–0.916] | 0.733 [0.631–0.828] | 0.932 [0.898–0.964] | 0.818 [0.723–0.906] | 0.773 [0.693–0.850] |
| | MERL-ViT (5.6 M) | 0.702 [0.654–0.747] | 0.605 [0.508–0.724] | 0.743 [0.691–0.805] | 0.495 [0.363–0.570] | 0.544 [0.424–0.615] |
| | MERL-ResNet (4.6 M) | 0.914 [0.887–0.947] | 0.802 [0.721–0.876] | 0.961 [0.936–0.985] | 0.896 [0.826–0.960] | 0.847 [0.794–0.901] |
| | ECG-CPC (2.9 M) | **0.918 [0.889–0.945]** | **0.837 [0.748–0.904]** | 0.952 [0.922–0.979] | 0.878 [0.802–0.944] | **0.857 [0.790–0.904]** |

**Training strategies**: FS = from scratch; LP = Linear Probing; FT = Full Fine-Tuning. All threshold-dependent metrics computed at a fixed decision threshold of 0.5. Values are mean [95% CI]. Bold indicates the best value per metric within each strategy, excluding models with degenerate (single-class) predictions.

Table 12: Patient-level secondary metrics on the BrSwiss-3% dataset (within-site).

| | Model | Accuracy | Sensitivity | Specificity | Precision | F1 |
|---|---|---|---|---|---|---|
| FS | HuBERTECG (94.6 M) | 0.706 [0.661–0.769] | 0.000 [0.000–0.000] | 1.000 [1.000–1.000] | 0.000 [0.000–0.000] | 0.000 [0.000–0.000] |
| | ECG-FM (92.1 M) | 0.706 [0.661–0.769] | 0.000 [0.000–0.000] | 1.000 [1.000–1.000] | 0.000 [0.000–0.000] | 0.000 [0.000–0.000] |
| | ST-MEM (90.3 M) | 0.706 [0.661–0.769] | 0.000 [0.000–0.000] | 1.000 [1.000–1.000] | 0.000 [0.000–0.000] | 0.000 [0.000–0.000] |
| | ECG-JEPA (87.1 M) | **0.825 [0.779–0.866]** | 0.535 [0.426–0.650] | 0.947 [0.924–0.975] | 0.807 [0.694–0.894] | **0.643 [0.539–0.730]** |
| | ECGFounder (33.8 M) | 0.733 [0.687–0.783] | **0.605 [0.518–0.721]** | 0.786 [0.746–0.827] | 0.542 [0.450–0.632] | 0.571 [0.501–0.651] |
| | ECGFM-KED (9.6 M) | 0.777 [0.726–0.834] | 0.291 [0.203–0.392] | **0.981 [0.962–0.996]** | 0.862 [0.762–0.965] | 0.435 [0.322–0.535] |
| | MERL-ViT (5.6 M) | 0.798 [0.750–0.843] | 0.360 [0.248–0.473] | 0.981 [0.960–0.995] | **0.886 [0.763–0.976]** | 0.512 [0.380–0.614] |
| | MERL-ResNet (4.6 M) | 0.777 [0.730–0.820] | 0.384 [0.263–0.493] | 0.942 [0.914–0.966] | 0.733 [0.578–0.839] | 0.504 [0.370–0.596] |
| | ECG-CPC (2.9 M) | 0.788 [0.745–0.829] | 0.512 [0.425–0.592] | 0.903 [0.871–0.936] | 0.688 [0.568–0.795] | 0.587 [0.500–0.665] |
| LP | HuBERTECG (94.6 M) | 0.798 [0.741–0.846] | 0.593 [0.486–0.700] | 0.883 [0.833–0.923] | 0.680 [0.566–0.780] | 0.633 [0.531–0.722] |
| | ECG-FM (92.1 M) | 0.843 [0.803–0.877] | 0.767 [0.691–0.853] | 0.874 [0.832–0.911] | 0.717 [0.621–0.802] | 0.742 [0.670–0.798] |
| | ST-MEM (90.3 M) | 0.843 [0.806–0.878] | 0.826 [0.765–0.893] | 0.850 [0.804–0.897] | 0.696 [0.580–0.779] | 0.755 [0.667–0.811] |
| | ECG-JEPA (87.1 M) | 0.805 [0.757–0.849] | 0.535 [0.442–0.622] | 0.917 [0.878–0.947] | 0.730 [0.578–0.820] | 0.617 [0.530–0.692] |
| | ECGFounder (33.8 M) | **0.873 [0.836–0.909]** | 0.756 [0.667–0.820] | 0.922 [0.884–0.958] | 0.802 [0.703–0.895] | **0.778 [0.714–0.836]** |
| | ECGFM-KED (9.6 M) | 0.483 [0.429–0.534] | **0.919 [0.847–0.965]** | 0.301 [0.251–0.377] | 0.354 [0.287–0.408] | 0.511 [0.436–0.565] |
| | MERL-ViT (5.6 M) | 0.801 [0.758–0.846] | 0.465 [0.340–0.566] | 0.942 [0.911–0.969] | 0.769 [0.639–0.864] | 0.580 [0.453–0.658] |
| | MERL-ResNet (4.6 M) | 0.671 [0.625–0.723] | 0.023 [0.000–0.058] | 0.942 [0.908–0.966] | 0.143 [0.000–0.265] | 0.040 [0.000–0.093] |
| | ECG-CPC (2.9 M) | 0.794 [0.750–0.839] | 0.337 [0.220–0.413] | **0.985 [0.970–1.000]** | **0.906 [0.779–1.000]** | 0.491 [0.347–0.571] |
| FT | HuBERTECG (94.6 M) | 0.812 [0.762–0.853] | 0.686 [0.605–0.774] | 0.864 [0.821–0.907] | 0.678 [0.562–0.767] | 0.682 [0.605–0.750] |
| | ECG-FM (92.1 M) | 0.723 [0.671–0.774] | 0.884 [0.811–0.944] | 0.655 [0.590–0.716] | 0.517 [0.411–0.600] | 0.652 [0.554–0.719] |
| | ST-MEM (90.3 M) | 0.846 [0.806–0.887] | **0.895 [0.846–0.939]** | 0.825 [0.772–0.879] | 0.681 [0.571–0.769] | 0.774 [0.685–0.833] |
| | ECG-JEPA (87.1 M) | 0.815 [0.769–0.855] | 0.488 [0.396–0.582] | 0.952 [0.921–0.975] | 0.808 [0.690–0.897] | 0.609 [0.519–0.691] |
| | ECGFounder (33.8 M) | 0.774 [0.724–0.822] | 0.872 [0.806–0.934] | 0.733 [0.672–0.785] | 0.577 [0.481–0.649] | 0.694 [0.613–0.761] |
| | ECGFM-KED (9.6 M) | 0.832 [0.787–0.879] | 0.558 [0.456–0.655] | 0.947 [0.921–0.974] | 0.814 [0.715–0.901] | 0.662 [0.563–0.754] |
| | MERL-ViT (5.6 M) | 0.815 [0.774–0.865] | 0.395 [0.297–0.519] | **0.990 [0.975–1.000]** | **0.944 [0.821–1.000]** | 0.557 [0.436–0.675] |
| | MERL-ResNet (4.6 M) | **0.880 [0.846–0.914]** | 0.709 [0.612–0.797] | 0.952 [0.925–0.981] | 0.859 [0.778–0.933] | **0.777 [0.700–0.833]** |
| | ECG-CPC (2.9 M) | 0.863 [0.819–0.901] | 0.639 [0.544–0.753] | 0.956 [0.931–0.978] | 0.859 [0.777–0.922] | 0.733 [0.658–0.811] |

**Training strategies**: FS = from scratch; LP = Linear Probing; FT = Full Fine-Tuning. All threshold-dependent metrics computed at a fixed decision threshold of 0.5. Values are mean [95% CI]. Bold indicates the best value per metric within each strategy, excluding models with degenerate (single-class) predictions.

Table 13: Patient-level secondary metrics on the HUCA dataset (within-site).

| | Model | Accuracy | Sensitivity | Specificity | Precision | F1 |
|---|---|---|---|---|---|---|
| FS | HuBERTECG (94.6 M) | 0.595 [0.541–0.657] | 0.329 [0.241–0.459] | 0.665 [0.605–0.720] | 0.207 [0.144–0.274] | 0.254 [0.185–0.339] |
| | ECG-FM (92.1 M) | 0.791 [0.749–0.838] | 0.000 [0.000–0.000] | 1.000 [1.000–1.000] | 0.000 [0.000–0.000] | 0.000 [0.000–0.000] |
| | ST-MEM (90.3 M) | 0.791 [0.749–0.838] | 0.000 [0.000–0.000] | 1.000 [1.000–1.000] | 0.000 [0.000–0.000] | 0.000 [0.000–0.000] |
| | ECG-JEPA (87.1 M) | 0.791 [0.749–0.838] | 0.000 [0.000–0.000] | 1.000 [1.000–1.000] | 0.000 [0.000–0.000] | 0.000 [0.000–0.000] |
| | ECGFounder (33.8 M) | **0.876 [0.840–0.906]** | **0.711 [0.619–0.812]** | 0.920 [0.884–0.950] | 0.701 [0.582–0.794] | **0.706 [0.609–0.777]** |
| | ECGFM-KED (9.6 M) | 0.849 [0.804–0.877] | 0.395 [0.279–0.503] | **0.969 [0.947–0.987]** | **0.769 [0.636–0.905]** | 0.522 [0.391–0.624] |
| | MERL-ViT (5.6 M) | 0.788 [0.748–0.833] | 0.000 [0.000–0.000] | 0.997 [0.990–1.000] | 0.000 [0.000–0.000] | 0.000 [0.000–0.000] |
| | MERL-ResNet (4.6 M) | 0.857 [0.814–0.890] | 0.500 [0.387–0.583] | 0.951 [0.915–0.974] | 0.731 [0.596–0.842] | 0.594 [0.471–0.661] |
| | ECG-CPC (2.9 M) | 0.860 [0.824–0.897] | 0.618 [0.504–0.705] | 0.923 [0.889–0.958] | 0.681 [0.551–0.813] | 0.648 [0.560–0.729] |
| LP | HuBERTECG (94.6 M) | 0.818 [0.780–0.858] | 0.276 [0.179–0.376] | 0.962 [0.943–0.985] | 0.656 [0.491–0.836] | 0.389 [0.272–0.489] |
| | ECG-FM (92.1 M) | 0.755 [0.716–0.793] | **0.763 [0.671–0.870]** | 0.753 [0.705–0.800] | 0.450 [0.365–0.542] | 0.566 [0.484–0.648] |
| | ST-MEM (90.3 M) | 0.860 [0.817–0.899] | 0.605 [0.500–0.730] | 0.927 [0.896–0.952] | 0.687 [0.592–0.788] | 0.643 [0.538–0.737] |
| | ECG-JEPA (87.1 M) | 0.862 [0.832–0.894] | 0.605 [0.494–0.706] | 0.930 [0.903–0.958] | 0.697 [0.586–0.794] | 0.648 [0.556–0.730] |
| | ECGFounder (33.8 M) | **0.873 [0.838–0.905]** | 0.618 [0.495–0.713] | 0.941 [0.910–0.970] | 0.734 [0.621–0.842] | **0.671 [0.570–0.750]** |
| | ECGFM-KED (9.6 M) | 0.785 [0.747–0.829] | 0.000 [0.000–0.000] | 0.993 [0.983–1.000] | 0.000 [0.000–0.000] | 0.000 [0.000–0.000] |
| | MERL-ViT (5.6 M) | 0.807 [0.760–0.850] | 0.395 [0.285–0.505] | 0.916 [0.886–0.941] | 0.556 [0.439–0.670] | 0.462 [0.360–0.570] |
| | MERL-ResNet (4.6 M) | 0.838 [0.807–0.882] | 0.697 [0.587–0.804] | 0.875 [0.836–0.917] | 0.596 [0.479–0.692] | 0.642 [0.556–0.726] |
| | ECG-CPC (2.9 M) | 0.851 [0.811–0.884] | 0.395 [0.274–0.500] | **0.972 [0.950–0.989]** | **0.789 [0.649–0.911]** | 0.526 [0.395–0.618] |
| FT | HuBERTECG (94.6 M) | 0.843 [0.801–0.884] | 0.605 [0.460–0.721] | 0.906 [0.868–0.943] | 0.630 [0.503–0.713] | 0.617 [0.482–0.703] |
| | ECG-FM (92.1 M) | 0.744 [0.696–0.784] | **0.816 [0.725–0.909]** | 0.725 [0.674–0.768] | 0.440 [0.347–0.513] | 0.571 [0.475–0.639] |
| | ST-MEM (90.3 M) | 0.865 [0.829–0.898] | 0.553 [0.441–0.653] | 0.948 [0.923–0.970] | 0.737 [0.617–0.829] | 0.632 [0.532–0.706] |
| | ECG-JEPA (87.1 M) | 0.879 [0.835–0.909] | 0.605 [0.464–0.709] | 0.951 [0.922–0.981] | 0.767 [0.637–0.886] | **0.676 [0.553–0.747]** |
| | ECGFounder (33.8 M) | 0.868 [0.826–0.898] | 0.645 [0.525–0.732] | 0.927 [0.891–0.962] | 0.700 [0.554–0.804] | 0.671 [0.552–0.748] |
| | ECGFM-KED (9.6 M) | 0.807 [0.769–0.846] | 0.145 [0.076–0.217] | **0.983 [0.965–0.996]** | 0.688 [0.411–0.923] | 0.239 [0.137–0.339] |
| | MERL-ViT (5.6 M) | 0.780 [0.737–0.820] | 0.421 [0.316–0.548] | 0.875 [0.836–0.910] | 0.471 [0.339–0.591] | 0.444 [0.341–0.539] |
| | MERL-ResNet (4.6 M) | **0.881 [0.850–0.910]** | 0.553 [0.446–0.660] | 0.969 [0.946–0.987] | **0.824 [0.717–0.914]** | 0.661 [0.563–0.748] |
| | ECG-CPC (2.9 M) | 0.873 [0.839–0.904] | 0.513 [0.419–0.618] | 0.969 [0.951–0.986] | 0.812 [0.685–0.914] | 0.629 [0.540–0.722] |

**Training strategies**: FS = from scratch; LP = Linear Probing; FT = Full Fine-Tuning. All threshold-dependent metrics computed at a fixed decision threshold of 0.5. Values are mean [95% CI]. Bold indicates the best value per metric within each strategy, excluding models with degenerate (single-class) predictions.

Table 14: Patient-level secondary metrics - train BrSwiss-100% and test on HUCA (cross-site).

| | Model | Accuracy | Sensitivity | Specificity | Precision | F1 |
|---|---|---|---|---|---|---|
| FS | HuBERTECG (94.6 M) | 0.791 [0.751–0.836] | 0.000 [0.000–0.000] | 1.000 [1.000–1.000] | 0.000 [0.000–0.000] | 0.000 [0.000–0.000] |
| | ECG-FM (92.1 M) | 0.791 [0.751–0.836] | 0.000 [0.000–0.000] | 1.000 [1.000–1.000] | 0.000 [0.000–0.000] | 0.000 [0.000–0.000] |
| | ST-MEM (90.3 M) | **0.771 [0.727–0.820]** | 0.013 [0.000–0.041] | **0.972 [0.951–0.989]** | 0.111 [0.000–0.333] | 0.024 [0.000–0.073] |
| | ECG-JEPA (87.1 M) | 0.529 [0.477–0.569] | **0.737 [0.597–0.835]** | 0.474 [0.415–0.518] | **0.271 [0.205–0.329]** | **0.396 [0.311–0.461]** |
| | ECGFounder (33.8 M) | 0.592 [0.540–0.641] | 0.474 [0.383–0.583] | 0.624 [0.572–0.684] | 0.250 [0.175–0.325] | 0.327 [0.246–0.408] |
| | ECGFM-KED (9.6 M) | 0.543 [0.487–0.588] | 0.605 [0.503–0.693] | 0.526 [0.464–0.577] | 0.253 [0.176–0.316] | 0.357 [0.264–0.428] |
| | MERL-ViT (5.6 M) | 0.559 [0.515–0.610] | 0.421 [0.309–0.524] | 0.596 [0.546–0.648] | 0.216 [0.149–0.270] | 0.286 [0.206–0.349] |
| | MERL-ResNet (4.6 M) | 0.485 [0.433–0.533] | 0.566 [0.452–0.664] | 0.463 [0.404–0.512] | 0.218 [0.147–0.274] | 0.315 [0.228–0.381] |
| | ECG-CPC (2.9 M) | 0.581 [0.526–0.638] | 0.513 [0.410–0.618] | 0.599 [0.547–0.660] | 0.253 [0.184–0.325] | 0.339 [0.261–0.416] |
| LP | HuBERTECG (94.6 M) | 0.209 [0.164–0.249] | 1.000 [1.000–1.000] | 0.000 [0.000–0.000] | 0.209 [0.164–0.249] | 0.346 [0.282–0.399] |
| | ECG-FM (92.1 M) | 0.463 [0.404–0.521] | 0.842 [0.750–0.933] | 0.362 [0.301–0.419] | **0.259 [0.203–0.312]** | **0.396 [0.327–0.461]** |
| | ST-MEM (90.3 M) | 0.455 [0.398–0.490] | 0.855 [0.776–0.926] | 0.348 [0.292–0.387] | 0.258 [0.197–0.301] | 0.396 [0.317–0.449] |
| | ECG-JEPA (87.1 M) | 0.510 [0.461–0.552] | 0.539 [0.449–0.660] | 0.502 [0.447–0.552] | 0.223 [0.170–0.279] | 0.315 [0.249–0.386] |
| | ECGFounder (33.8 M) | 0.355 [0.313–0.396] | 0.711 [0.614–0.796] | 0.261 [0.215–0.307] | 0.203 [0.155–0.250] | 0.316 [0.253–0.375] |
| | ECGFM-KED (9.6 M) | 0.240 [0.196–0.291] | **0.908 [0.837–0.965]** | 0.063 [0.037–0.091] | 0.204 [0.161–0.249] | 0.333 [0.274–0.394] |
| | MERL-ViT (5.6 M) | 0.576 [0.537–0.628] | 0.368 [0.269–0.454] | 0.631 [0.581–0.678] | 0.209 [0.137–0.283] | 0.267 [0.182–0.338] |
| | MERL-ResNet (4.6 M) | **0.656 [0.610–0.703]** | 0.118 [0.043–0.194] | **0.798 [0.756–0.841]** | 0.134 [0.048–0.216] | 0.126 [0.046–0.196] |
| | ECG-CPC (2.9 M) | 0.595 [0.537–0.646] | 0.368 [0.275–0.476] | 0.655 [0.593–0.704] | 0.221 [0.145–0.289] | 0.276 [0.201–0.348] |
| FT | HuBERTECG (94.6 M) | 0.209 [0.164–0.249] | 1.000 [1.000–1.000] | 0.000 [0.000–0.000] | 0.209 [0.164–0.249] | 0.346 [0.282–0.399] |
| | ECG-FM (92.1 M) | 0.449 [0.399–0.503] | 0.737 [0.650–0.820] | 0.373 [0.320–0.434] | 0.237 [0.190–0.293] | 0.359 [0.299–0.425] |
| | ST-MEM (90.3 M) | 0.460 [0.402–0.500] | 0.855 [0.776–0.926] | 0.355 [0.290–0.394] | **0.260 [0.198–0.306]** | **0.399 [0.319–0.452]** |
| | ECG-JEPA (87.1 M) | 0.452 [0.412–0.499] | 0.632 [0.522–0.752] | 0.404 [0.352–0.460] | 0.219 [0.166–0.281] | 0.325 [0.260–0.402] |
| | ECGFounder (33.8 M) | 0.499 [0.453–0.546] | 0.750 [0.679–0.844] | 0.432 [0.384–0.483] | 0.259 [0.189–0.317] | 0.385 [0.297–0.453] |
| | ECGFM-KED (9.6 M) | 0.537 [0.483–0.580] | 0.579 [0.460–0.669] | 0.526 [0.476–0.570] | 0.244 [0.176–0.307] | 0.344 [0.258–0.412] |
| | MERL-ViT (5.6 M) | **0.548 [0.507–0.597]** | 0.342 [0.254–0.437] | **0.603 [0.554–0.649]** | 0.186 [0.125–0.238] | 0.241 [0.168–0.300] |
| | MERL-ResNet (4.6 M) | 0.339 [0.278–0.387] | 0.895 [0.800–0.957] | 0.192 [0.155–0.238] | 0.227 [0.169–0.274] | 0.362 [0.282–0.423] |
| | ECG-CPC (2.9 M) | 0.397 [0.344–0.444] | **0.934 [0.885–0.988]** | 0.254 [0.212–0.297] | 0.249 [0.190–0.295] | 0.393 [0.314–0.453] |

**Training strategies**: FS = from scratch; LP = Linear Probing; FT = Full Fine-Tuning. All threshold-dependent metrics computed at a fixed decision threshold of 0.5. Values are mean [95% CI]. Bold indicates the best value per metric within each strategy, excluding models with degenerate (single-class) predictions.

Table 15: Patient-level secondary metrics - train BrSwiss-3% and test on HUCA (cross-site).

| | Model | Accuracy | Sensitivity | Specificity | Precision | F1 |
|---|---|---|---|---|---|---|
| FS | HuBERTECG (94.6 M) | 0.791 [0.751–0.836] | 0.000 [0.000–0.000] | 1.000 [1.000–1.000] | 0.000 [0.000–0.000] | 0.000 [0.000–0.000] |
| | ECG-FM (92.1 M) | 0.791 [0.751–0.836] | 0.000 [0.000–0.000] | 1.000 [1.000–1.000] | 0.000 [0.000–0.000] | 0.000 [0.000–0.000] |
| | ST-MEM (90.3 M) | 0.791 [0.751–0.836] | 0.000 [0.000–0.000] | 1.000 [1.000–1.000] | 0.000 [0.000–0.000] | 0.000 [0.000–0.000] |
| | ECG-JEPA (87.1 M) | 0.463 [0.419–0.501] | **0.737 [0.611–0.846]** | 0.390 [0.353–0.434] | **0.242 [0.184–0.297]** | **0.365 [0.291–0.433]** |
| | ECGFounder (33.8 M) | 0.650 [0.607–0.700] | 0.224 [0.154–0.320] | 0.763 [0.720–0.803] | 0.200 [0.138–0.285] | 0.211 [0.154–0.292] |
| | ECGFM-KED (9.6 M) | 0.485 [0.442–0.536] | 0.632 [0.525–0.713] | 0.446 [0.399–0.503] | 0.232 [0.168–0.291] | 0.339 [0.257–0.406] |
| | MERL-ViT (5.6 M) | 0.556 [0.508–0.598] | 0.368 [0.269–0.460] | 0.606 [0.555–0.657] | 0.199 [0.133–0.264] | 0.258 [0.178–0.323] |
| | MERL-ResNet (4.6 M) | **0.736 [0.687–0.781]** | 0.053 [0.013–0.102] | **0.916 [0.881–0.944]** | 0.143 [0.032–0.254] | 0.077 [0.019–0.141] |
| | ECG-CPC (2.9 M) | 0.730 [0.686–0.788] | 0.105 [0.042–0.173] | 0.895 [0.858–0.928] | 0.210 [0.092–0.333] | 0.140 [0.057–0.223] |
| LP | HuBERTECG (94.6 M) | 0.209 [0.164–0.249] | 1.000 [1.000–1.000] | 0.000 [0.000–0.000] | 0.209 [0.164–0.249] | 0.346 [0.282–0.399] |
| | ECG-FM (92.1 M) | 0.490 [0.445–0.540] | 0.895 [0.834–0.964] | 0.383 [0.316–0.432] | 0.278 [0.219–0.334] | **0.424 [0.352–0.490]** |
| | ST-MEM (90.3 M) | 0.273 [0.220–0.324] | **0.947 [0.890–1.000]** | 0.094 [0.061–0.128] | 0.217 [0.166–0.258] | 0.353 [0.284–0.406] |
| | ECG-JEPA (87.1 M) | **0.680 [0.629–0.734]** | 0.329 [0.241–0.417] | 0.773 [0.720–0.821] | **0.278 [0.201–0.365]** | 0.301 [0.221–0.377] |
| | ECGFounder (33.8 M) | 0.457 [0.408–0.497] | 0.618 [0.512–0.705] | 0.415 [0.356–0.468] | 0.219 [0.164–0.267] | 0.323 [0.250–0.376] |
| | ECGFM-KED (9.6 M) | 0.242 [0.193–0.288] | 0.947 [0.879–0.987] | 0.056 [0.029–0.084] | 0.210 [0.164–0.253] | 0.344 [0.278–0.402] |
| | MERL-ViT (5.6 M) | 0.590 [0.541–0.635] | 0.250 [0.176–0.344] | 0.679 [0.635–0.720] | 0.171 [0.108–0.237] | 0.203 [0.141–0.284] |
| | MERL-ResNet (4.6 M) | 0.656 [0.610–0.703] | 0.118 [0.043–0.194] | **0.798 [0.756–0.841]** | 0.134 [0.048–0.216] | 0.126 [0.046–0.196] |
| | ECG-CPC (2.9 M) | 0.606 [0.547–0.660] | 0.355 [0.266–0.459] | 0.672 [0.610–0.715] | 0.223 [0.150–0.289] | 0.274 [0.200–0.350] |
| FT | HuBERTECG (94.6 M) | 0.209 [0.164–0.249] | 1.000 [1.000–1.000] | 0.000 [0.000–0.000] | 0.209 [0.164–0.249] | 0.346 [0.282–0.399] |
| | ECG-FM (92.1 M) | 0.438 [0.392–0.482] | 0.829 [0.755–0.895] | 0.335 [0.290–0.382] | 0.248 [0.197–0.298] | **0.382 [0.318–0.442]** |
| | ST-MEM (90.3 M) | 0.240 [0.191–0.278] | **0.921 [0.859–0.976]** | 0.059 [0.030–0.087] | 0.206 [0.158–0.243] | 0.337 [0.271–0.386] |
| | ECG-JEPA (87.1 M) | **0.669 [0.621–0.721]** | 0.303 [0.208–0.394] | **0.767 [0.721–0.811]** | **0.256 [0.173–0.338]** | 0.277 [0.197–0.346] |
| | ECGFounder (33.8 M) | 0.303 [0.255–0.338] | 0.842 [0.777–0.919] | 0.160 [0.120–0.191] | 0.210 [0.166–0.249] | 0.336 [0.278–0.388] |
| | ECGFM-KED (9.6 M) | 0.614 [0.573–0.668] | 0.355 [0.259–0.454] | 0.683 [0.638–0.734] | 0.229 [0.153–0.302] | 0.278 [0.198–0.365] |
| | MERL-ViT (5.6 M) | 0.510 [0.464–0.554] | 0.342 [0.247–0.427] | 0.554 [0.506–0.599] | 0.169 [0.105–0.213] | 0.226 [0.146–0.276] |
| | MERL-ResNet (4.6 M) | 0.408 [0.358–0.448] | 0.842 [0.732–0.934] | 0.293 [0.258–0.333] | 0.240 [0.175–0.285] | 0.373 [0.285–0.425] |
| | ECG-CPC (2.9 M) | 0.653 [0.596–0.704] | 0.237 [0.151–0.340] | 0.763 [0.700–0.802] | 0.209 [0.136–0.298] | 0.222 [0.151–0.303] |

**Training strategies**: FS = from scratch; LP = Linear Probing; FT = Full Fine-Tuning. All threshold-dependent metrics computed at a fixed decision threshold of 0.5. Values are mean [95% CI]. Bold indicates the best value per metric within each strategy, excluding models with degenerate (single-class) predictions.

Table 16: Patient-level secondary metrics - train HUCA and test on BrSwiss (cross-site).

| | Model | Accuracy | Sensitivity | Specificity | Precision | F1 |
|---|---|---|---|---|---|---|
| FS | HuBERTECG (94.6 M) | 0.294 [0.248–0.344] | 1.000 [1.000–1.000] | 0.000 [0.000–0.000] | 0.294 [0.248–0.344] | 0.455 [0.398–0.512] |
| | ECG-FM (92.1 M) | 0.706 [0.656–0.752] | 0.000 [0.000–0.000] | 1.000 [1.000–1.000] | 0.000 [0.000–0.000] | 0.000 [0.000–0.000] |
| | ST-MEM (90.3 M) | 0.706 [0.656–0.752] | 0.000 [0.000–0.000] | 1.000 [1.000–1.000] | 0.000 [0.000–0.000] | 0.000 [0.000–0.000] |
| | ECG-JEPA (87.1 M) | 0.706 [0.656–0.752] | 0.000 [0.000–0.000] | 1.000 [1.000–1.000] | 0.000 [0.000–0.000] | 0.000 [0.000–0.000] |
| | ECGFounder (33.8 M) | 0.706 [0.645–0.753] | 0.116 [0.059–0.174] | 0.952 [0.924–0.981] | 0.500 [0.259–0.747] | 0.189 [0.098–0.272] |
| | ECGFM-KED (9.6 M) | 0.709 [0.656–0.759] | 0.058 [0.021–0.112] | 0.981 [0.962–0.995] | 0.556 [0.261–0.800] | 0.105 [0.040–0.194] |
| | MERL-ViT (5.6 M) | 0.712 [0.663–0.757] | 0.023 [0.000–0.053] | **1.000 [1.000–1.000]** | **1.000 [0.000–1.000]** | 0.045 [0.000–0.101] |
| | MERL-ResNet (4.6 M) | 0.712 [0.661–0.760] | 0.046 [0.011–0.095] | 0.990 [0.973–1.000] | 0.667 [0.161–1.000] | 0.087 [0.020–0.172] |
| | ECG-CPC (2.9 M) | **0.730 [0.666–0.772]** | **0.128 [0.062–0.192]** | 0.981 [0.958–0.995] | 0.733 [0.456–0.929] | **0.218 [0.112–0.307]** |
| LP | HuBERTECG (94.6 M) | 0.294 [0.248–0.344] | 1.000 [1.000–1.000] | 0.000 [0.000–0.000] | 0.294 [0.248–0.344] | 0.455 [0.398–0.512] |
| | ECG-FM (92.1 M) | 0.657 [0.596–0.714] | 0.326 [0.230–0.415] | 0.796 [0.741–0.842] | 0.400 [0.285–0.518] | 0.359 [0.262–0.446] |
| | ST-MEM (90.3 M) | 0.637 [0.570–0.683] | 0.174 [0.082–0.257] | 0.830 [0.769–0.873] | 0.300 [0.154–0.415] | 0.221 [0.112–0.306] |
| | ECG-JEPA (87.1 M) | 0.685 [0.623–0.737] | 0.012 [0.000–0.036] | 0.966 [0.936–0.987] | 0.125 [0.000–0.400] | 0.021 [0.000–0.067] |
| | ECGFounder (33.8 M) | 0.688 [0.630–0.737] | 0.046 [0.011–0.096] | 0.956 [0.931–0.985] | 0.308 [0.071–0.620] | 0.081 [0.020–0.162] |
| | ECGFM-KED (9.6 M) | **0.723 [0.664–0.776]** | 0.105 [0.037–0.168] | **0.981 [0.963–0.995]** | **0.692 [0.353–0.905]** | 0.182 [0.066–0.281] |
| | MERL-ViT (5.6 M) | 0.709 [0.642–0.764] | 0.198 [0.122–0.299] | 0.922 [0.883–0.957] | 0.515 [0.352–0.705] | 0.286 [0.182–0.408] |
| | MERL-ResNet (4.6 M) | 0.685 [0.634–0.731] | **0.477 [0.365–0.582]** | 0.772 [0.708–0.823] | 0.466 [0.364–0.568] | **0.471 [0.383–0.556]** |
| | ECG-CPC (2.9 M) | 0.723 [0.669–0.767] | 0.140 [0.068–0.211] | 0.966 [0.942–0.990] | 0.632 [0.444–0.862] | 0.229 [0.116–0.329] |
| FT | HuBERTECG (94.6 M) | 0.294 [0.248–0.344] | 1.000 [1.000–1.000] | 0.000 [0.000–0.000] | 0.294 [0.248–0.344] | 0.455 [0.398–0.512] |
| | ECG-FM (92.1 M) | **0.774 [0.733–0.827]** | **0.430 [0.339–0.535]** | 0.917 [0.888–0.952] | 0.685 [0.571–0.805] | **0.529 [0.431–0.615]** |
| | ST-MEM (90.3 M) | 0.575 [0.517–0.618] | 0.058 [0.005–0.130] | 0.791 [0.738–0.829] | 0.104 [0.009–0.227] | 0.075 [0.006–0.157] |
| | ECG-JEPA (87.1 M) | 0.712 [0.664–0.762] | 0.105 [0.047–0.172] | 0.966 [0.938–0.985] | 0.562 [0.258–0.744] | 0.176 [0.080–0.273] |
| | ECGFounder (33.8 M) | 0.671 [0.608–0.723] | 0.140 [0.066–0.206] | 0.893 [0.845–0.932] | 0.353 [0.180–0.508] | 0.200 [0.099–0.282] |
| | ECGFM-KED (9.6 M) | 0.706 [0.656–0.752] | 0.000 [0.000–0.000] | 1.000 [1.000–1.000] | 0.000 [0.000–0.000] | 0.000 [0.000–0.000] |
| | MERL-ViT (5.6 M) | 0.692 [0.628–0.750] | 0.256 [0.169–0.364] | 0.874 [0.835–0.917] | 0.458 [0.354–0.637] | 0.328 [0.229–0.445] |
| | MERL-ResNet (4.6 M) | 0.702 [0.642–0.747] | 0.105 [0.047–0.164] | 0.952 [0.914–0.975] | 0.474 [0.250–0.676] | 0.171 [0.081–0.255] |
| | ECG-CPC (2.9 M) | 0.719 [0.666–0.765] | 0.070 [0.016–0.121] | **0.990 [0.978–1.000]** | **0.750 [0.448–1.000]** | 0.128 [0.031–0.210] |

**Training strategies**: FS = from scratch; LP = Linear Probing; FT = Full Fine-Tuning. All threshold-dependent metrics computed at a fixed decision threshold of 0.5. Values are mean [95% CI]. Bold indicates the best value per metric within each strategy, excluding models with degenerate (single-class) predictions.

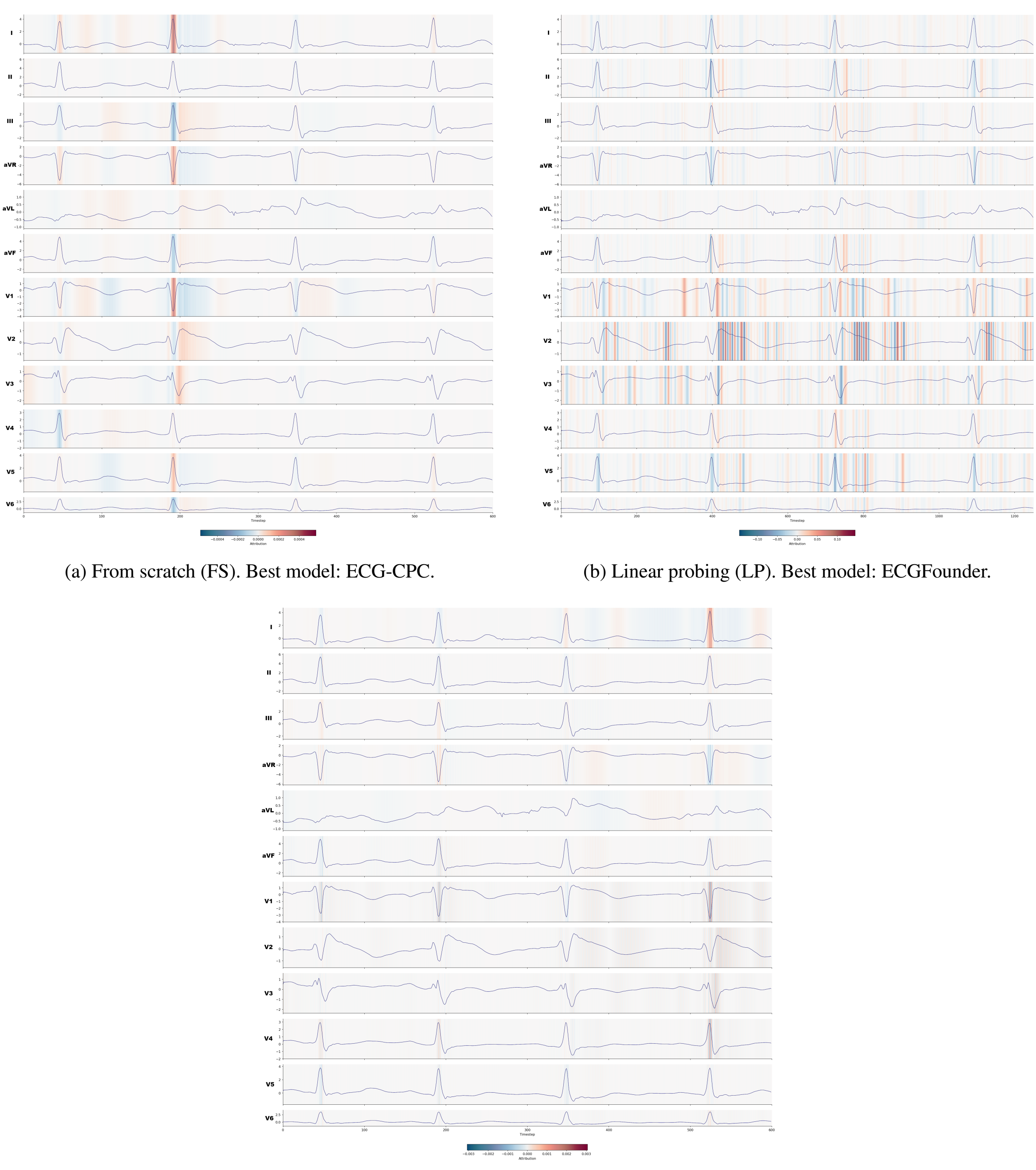

(a) From scratch (FS). Best model: ECG-CPC.

(b) Linear probing (LP). Best model: ECGFounder.

(c) Full fine-tuning (FT). Best model: ECG-CPC.

**Figure 5:** Integrated Gradients saliency maps on the BrSwiss-100% dataset for a representative Brugada syndrome patient, under the three adaptation strategies. Each panel shows all 12 ECG leads with the attribution heatmap overlaid. Each map is normalised by its own maximum absolute value.

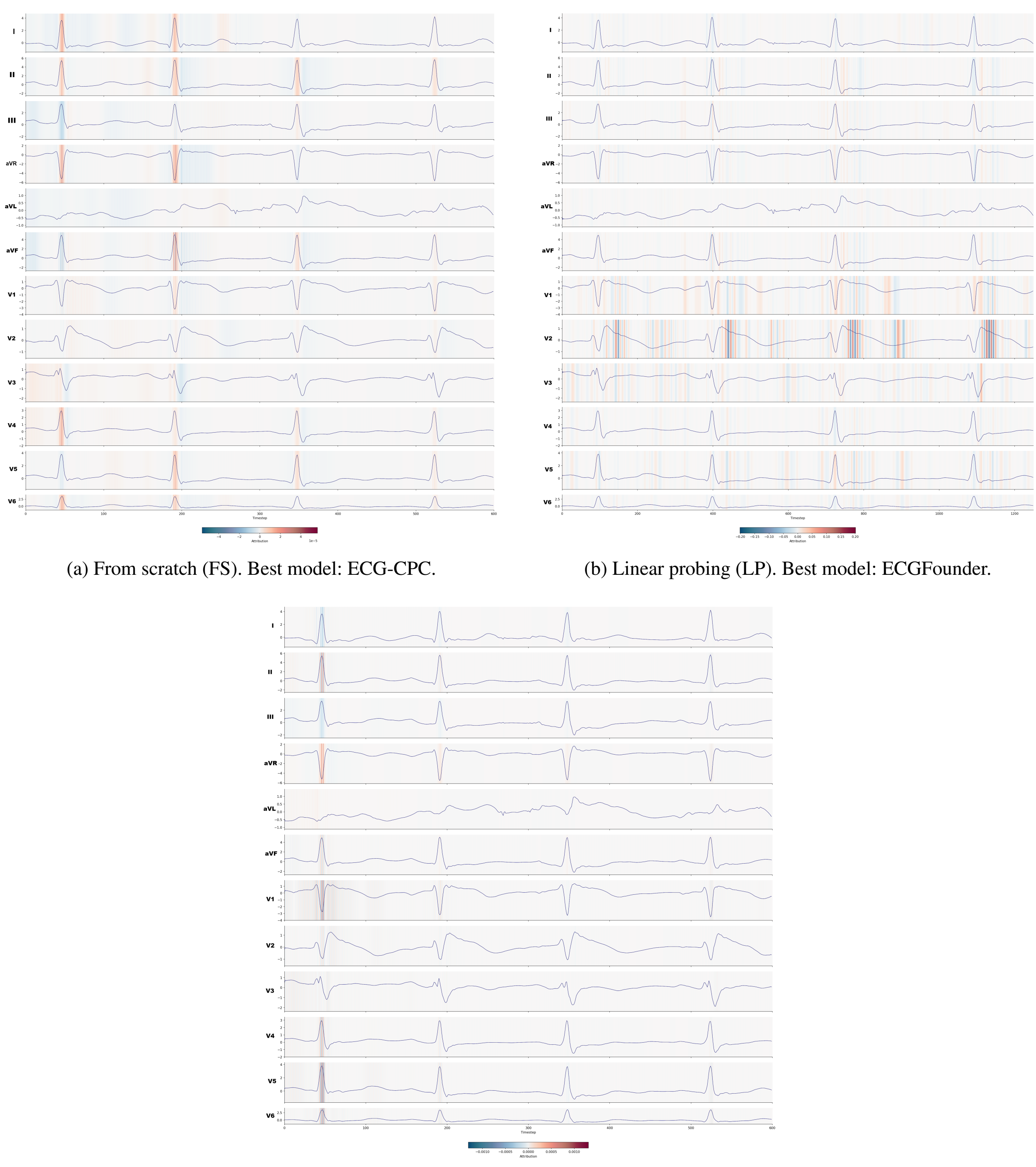

(a) From scratch (FS). Best model: ECG-CPC.

(b) Linear probing (LP). Best model: ECGFounder.

(c) Full fine-tuning (FT). Best model: ECG-CPC.

**Figure 6:** Integrated Gradients saliency maps on the BrSwiss-3% dataset for a representative Brugada syndrome patient, under the three adaptation strategies. Each panel shows all 12 ECG leads with the attribution heatmap overlaid. Each map is normalised by its own maximum absolute value.

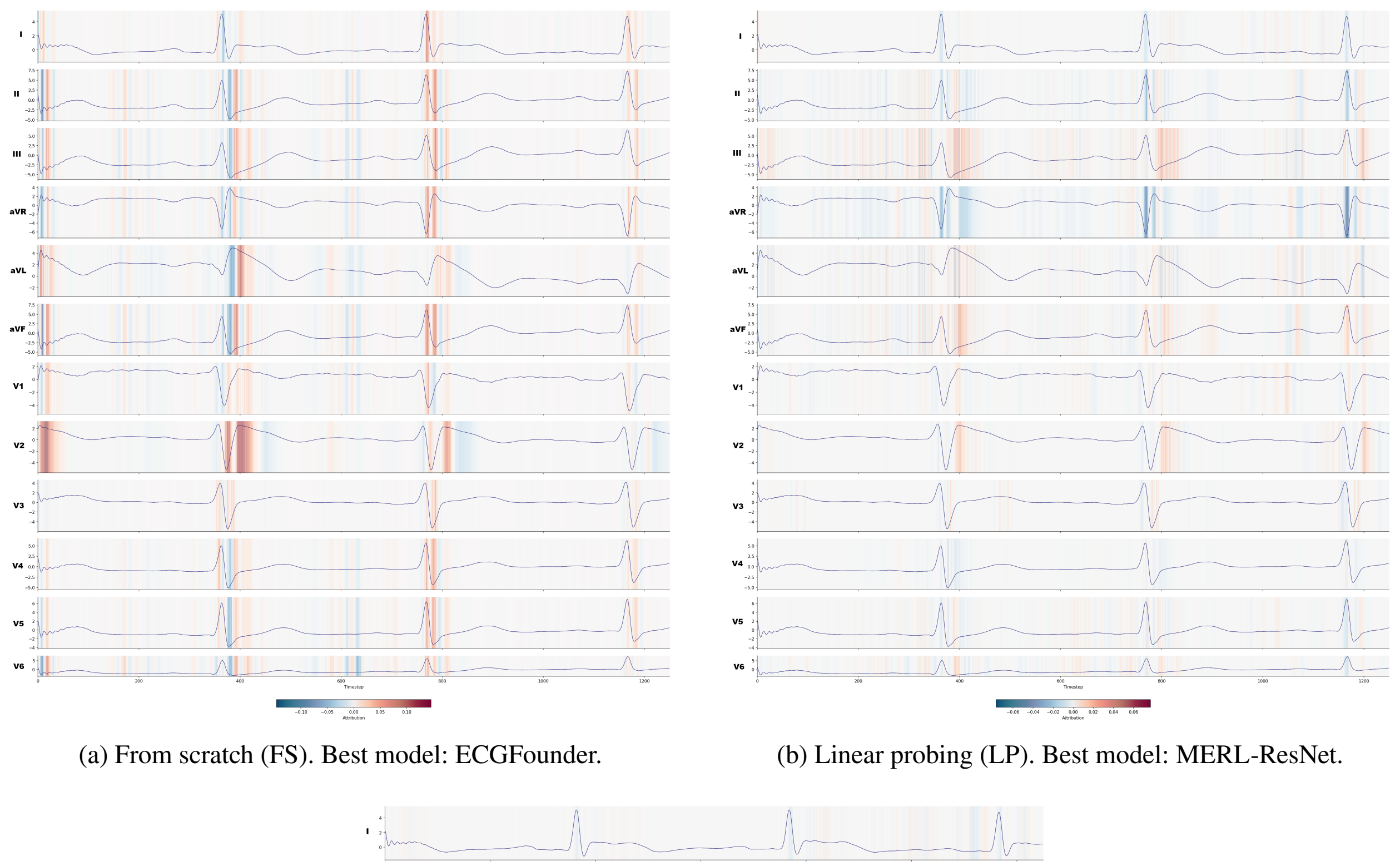

(a) From scratch (FS). Best model: ECGFounder.

(b) Linear probing (LP). Best model: MERL-ResNet.

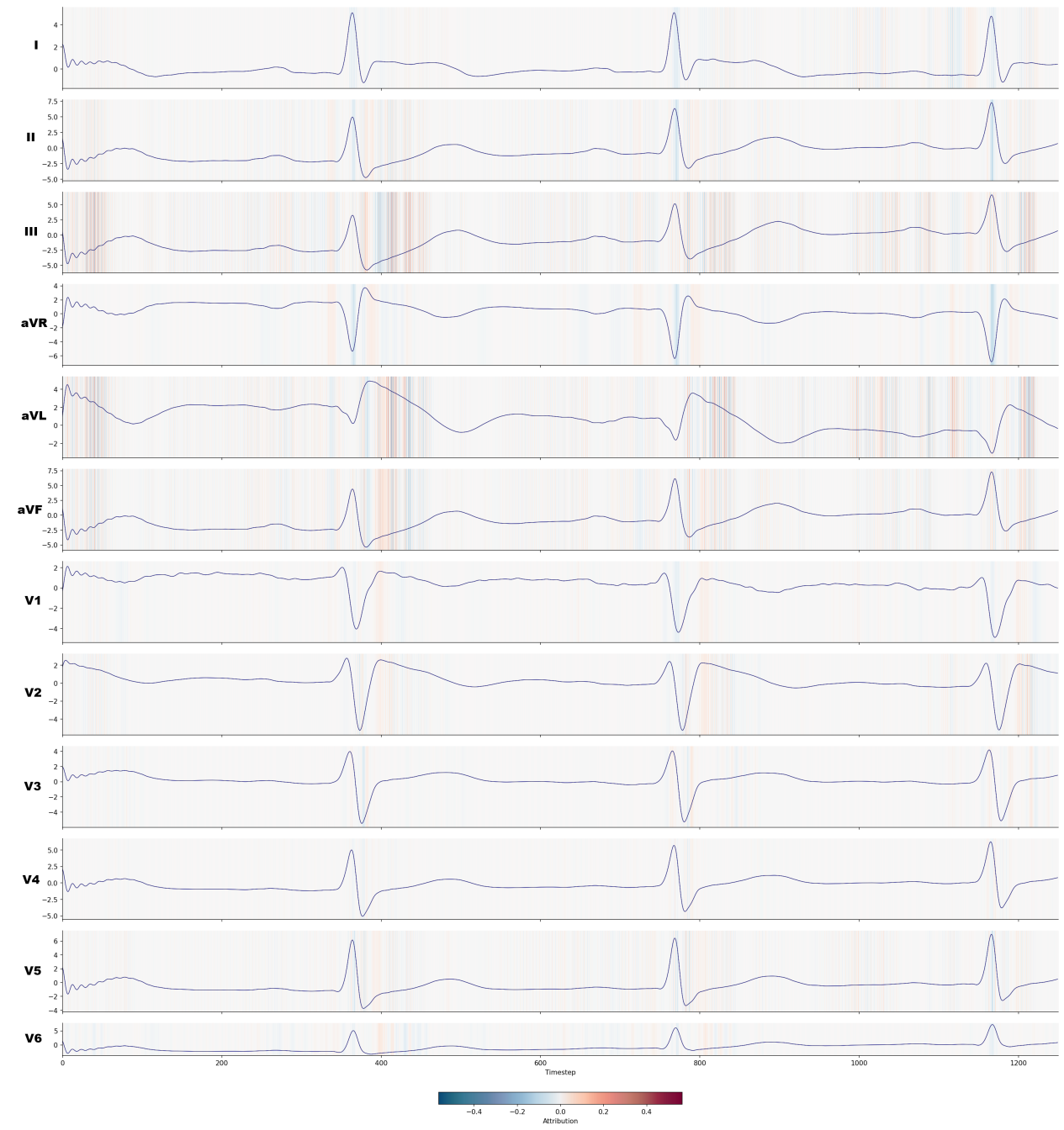

(c) Full fine-tuning (FT). Best model: MERL-ResNet.

**Figure 7:** Integrated Gradients saliency maps on the HUCA dataset for a representative Brugada syndrome patient, under the three adaptation strategies. Each panel shows all 12 ECG leads with the attribution heatmap overlaid. Each map is normalised by its own maximum absolute value.